\pdfoutput=1

\documentclass[11pt]{article}

\usepackage[]{acl}

\usepackage{times}
\usepackage{latexsym}

\usepackage[T1]{fontenc}

\usepackage[utf8]{inputenc}
\usepackage{times}
\usepackage{latexsym}
\usepackage{url}
\usepackage{enumitem}
\usepackage{amsmath,amssymb,amsfonts}
\usepackage{algorithmic}
\usepackage{caption}
\usepackage{subcaption}
\usepackage{graphicx}
\usepackage{textcomp}
\usepackage{url}
\usepackage{makecell}
\usepackage{multirow}
\usepackage{booktabs}
\usepackage{array}
\usepackage{microtype}
\usepackage{kotex}
\usepackage{array}
\usepackage{float}
\usepackage{booktabs}
\usepackage{hyperref}       
\usepackage{amsfonts}       
\usepackage{nicefrac}       
\usepackage{xcolor}         

\usepackage{kotex}
\usepackage{algorithm}
\usepackage{algorithmic}
\usepackage{todonotes}
\usepackage{multirow}
\usepackage{float}
\usepackage{comment}
\usepackage{todonotes}
\usepackage{booktabs, makecell, tabularx}

\newcolumntype{L}{>{\raggedright\arraybackslash}X}
\usepackage{siunitx}

\newcommand{\thickhline}{%
    \noalign {\ifnum 0=`}\fi \hrule height 1pt
    \futurelet \reserved@a \@xhline
}

\usepackage{comment}
\usepackage{todonotes}
\usepackage{xcolor}

\usepackage{microtype}

\makeatletter
\renewcommand*{\@fnsymbol}[1]{\ensuremath{\ifcase#1\or \dagger\or *\or \ddagger\or
   \mathsection\or \mathparagraph\or \|\or **\or \dagger\dagger
   \or \ddagger\ddagger \else\@ctrerr\fi}}
\makeatother

\title{Empirical Analysis of Korean Public AI Hub Parallel Corpora and in-depth Analysis using LIWC}

\author{Chanjun Park $^{1}$\thanks{\hspace*{0.5em}{These authors contributed equally to this work}} , Midan Shim $^{2\dagger}$, Sugyeong Eo $^{1\dagger}$, Seolhwa Lee $^{1\dagger}$, \\\textbf{Jaehyung Seo $^{1}$, Hyeonseok Moon $^{1}$, Heuiseok Lim$^{1}$ \thanks{\hspace*{0.5em}{Corresponding author.}}}\\
\\
  $^1$Korea University, $^2$Kyunghee University\\
  \texttt{\{bcj1210, djtnrud, whiteldark, seojae777, glee889, limhseok\}@korea.ac.kr}
  \\
  \texttt{hihello0426@gmail.com}
  \\
  }

\begin{document}
\maketitle
\begin{abstract}
Machine translation (MT) system aims to translate source language into target language. Recent studies on MT systems mainly focus on neural machine translation (NMT). One factor that significantly affects the performance of NMT is the availability of high-quality parallel corpora. However, high-quality parallel corpora concerning Korean are relatively scarce compared to those associated with other high-resource languages, such as German or Italian. To address this problem, AI Hub recently released seven types of parallel corpora for Korean. In this study, we conduct an in-depth verification of the quality of corresponding parallel corpora through Linguistic Inquiry and Word Count (LIWC) and several relevant experiments. LIWC is a word-counting software program that can analyze corpora in multiple ways and extract linguistic features as a dictionary base. To the best of our knowledge, this study is the first to use LIWC to analyze parallel corpora in the field of NMT. Our findings suggest the direction of further research toward obtaining the improved quality parallel corpora through our correlation analysis in LIWC and NMT performance.
\end{abstract}

\section{Introduction}

In recent years, the demand for machine translation (MT) systems has been continuously increasing and its importance is growing, especially for the industrial services. \cite{vieira2021understanding, zheng2019testing} Companies, such as Google, Facebook, Microsoft, Amazon, and Unbabel continue to conduct research and formulate plans to commercialize applications related to MT.

From the late 1950s, numerous MT-related projects were proceeded by mainly focusing on rule-based and statistical-based approaches before the advent of deep learning technology. As deep learning based neural machine translation (NMT) was proposed and adopted to several researches, it has been gradually figured out that more superior performance can be derived through NMT approach \cite{bahdanau2014neural,vaswani2017attention,lample2019cross,song2019mass}.

Followed by the adoption of deep learning based technique, the improvements of computing power (\emph{e.g. GPU}) and corresponding enhancement of parallel processing accelerated the advancement of NMT. Recently, release of open source frameworks, such as Pytorch\cite{NEURIPS2019_9015}, and lowered accessibility to the big data further facilitated vigorous and diverse research.

However, several issues considering the enhancement of the NMT system remain still. Representatively, limitations in ensuring the quality of data is an unresolved issue. As have previously been studied, the quality of the training data is deeply related to the NMT performance \cite{park2020toward,park2021study}. The major problem is that the process of building a high-quality parallel corpus is time-consuming and expensive, and it is significantly difficult for low-resource languages, such as Korean. Although data-augmentation techniques, such as back translation \cite{edunov2018understanding} and copied translation \cite{currey2017copied} have been introduced, as the human supervision is generally minimized or excluded in the data generation process, the quality of such pseudo-generated parallel corpus cannot be guaranteed \cite{burlot2019using, epaliyana2021improving}. This restricted the usage of pseudo-generated parallel to complements of human-labeled gold parallel corpus, rather than its substitutes \cite{imankulova2017improving}.

For the alleviation of above limitations, numerous studies on the collection of high-quality training data have been conducted, such as parallel corpus filtering (PCF) research and Data Dam project. PCF refers to a research field that aims to filter out low-quality noisy data (\emph{i.e.} sentence pairs) residing in the parallel corpus, and improve the overall quality of the corpus. PCF is currently being applied to various NMT studies and contributed to the advancement of the NMT systems \cite{koehn2019findings, park2020quality}. While the amount of training data caused significant impact on the statistical-based MT approaches, the quality of data is treated as more important than the amount of data in general deep learning-based MT approaches \cite{khayrallah2018impact, koehn-EtAl:2020:WMT}. Moreover, Data Dam \footnote{\url{http://www.data-alliance.kr/default/}} projects for building high-quality parallel corpora nationally are in progress. In the Republic of Korea, a large number of parallel corpora is open to the public through AI-Hub \footnote{\url{http://aihub.or.kr/}}, which is organized by the National Information Society Agency (NIA) \cite{park2020study}.

Following these research trends, where the quality is treated more importantly than the quantity in the data construction process, we analyzed the above Korean-English parallel corpus distributed by AI-Hub. Despite its sufficient amount of data, the quality of corresponding corpus has not been confirmed clearly. This may restrict the unconstrained utilization of such corpus in adoption to the NMT model, as low quality data may degrade the overall performance. 
In this study, we conducted several quality verification experiments including Linguistic Inquiry and Word Count (LIWC) \cite{pennebaker2001linguistic,tausczik2010psychological}, and clarified the quality and characteristics of such corpus. By analyzing various factors that can affect NMT performance, we proposed a method that can be applied in future research using the analysis results.

LIWC is a text-analysis tool that automatically analyzes the number of words in a sentence and classifies words with similar meanings and sentimental characteristics. LIWC extracts various interpersonal variables related to clinical, social, physiological, cognitive, psychological, and developmental contexts that cannot be detected using previous text-analysis programs. Additionally, LIWC comprises a variety of features for analyzing text. LIWC generally used to recognize linguistic markers for mental health study in Psychopathology such as detecting Narcissism\cite{holtzman2019linguistic}, schizophrenia\cite{bae2021schizophrenia}, bipolar disorder\cite{sekulic2018not}. However, LIWC provides various linguistic features, word count, gender bias and so on, so it can be used for various analyses. In this study, we use LIWC to analyze parallel corpora based on diverse properties. It is also first time to analyze corpus using LIWC.


In addition, we conduct baseline translation experiments by training transformer-base model structure \cite{vaswani2017attention} through all the parallel corpora given by AIhub. By analyzing MT performance of corresponding models, we propose further research directions on MT for the Korean language. The contributions of this study are as follows:

\begin{itemize}
\item{For the first time, we conduct a deep data analysis on AI-Hub data. To the best of our knowledge, this is the first time LIWC has been used to analyze corpora. This study acts as a milestone for further studies on NMT with respect to the Korean language.}

\item{We conduct baseline translation experiments on all the data in the AI-Hub parallel corpus. Our experiments provide a foundation for further research on Korean-based NMT.}

\item{We discovered that many factors might cause decreasing model performance, and we provide the direction that those factors could be filtered through our correlation analysis between LIWC and model performance.} 
\end{itemize}

\section{Related works and Background}
\subsection{Machine Translation}
Machine Translation (MT) refers to a computer system that translates source sentences into target sentences and has achieved significant performance improvements with the advent of deep learning. In 1951, Yehoshua-Bar-hillel first started research on MT at MIT \cite{kasher2012language}, and it has gradually been developed in the order of rule-based, statistical-based, and deep learning-based MT.

~\\
\textbf{Rule-based Machine Translation}
Rule-based MT (RBMT) \cite{dugast2007statistical,forcada2011apertium} is a translation method based on linguistics rules established by linguists, as well as traditional natural language processing such as lexical analysis, syntax analysis, and semantic analysis. For example, the Korean-English RBMT is a methodology that transfers Korean sentence in accordance with the English grammatical rules based on a process of morphological analysis and synthetic analysis, and translates Korean sentence as the source language into English sentence as the target language. This method has the advantage of conducting ideal translation of sentences that conform to the rules, but has the disadvantage of difficulty in extracting grammatical rules and requiring a lot of linguistic knowledge. It is also difficult to expand the translatable language pairs and numerous rules should be considered.

~\\
\textbf{Statistical Machine Translation}
Statistical MT (SMT) \cite{zens2002phrase,koehn2009statistical} is a method of translating using statistical prior knowledge learned from large scale parallel corpus. This method utilizes the alignment and co-occurrence based on statistical information between words from large scale parallel corpus.

SMT contains a translation, reordering and language model. It extracts the alignment of the source sentence and target sentence through the translation model, and predicts the probability of the target sentence through the language model. Unlike RBMT, this methodology can be developed without linguistic knowledge and generally higher performance can be obtained by increasing the amount of data. However, building large amounts of data is a challenging task and the context is difficult to understand, because the translation is carried out in words or phrases basis.

In the case of SMT, the methodology has changed according to the unit of translation. At the beginning of the study, translation was performed in words. However, in 2003, a translation method of multiple word bundles (i.e. phrase units), was proposed and showed better performance better than word units. The introduction of the concept of variables within the phrase is referred to "Hierarchical Phrase-Based SMT", which does not indicate a specific word, such as "eat an bread", but rather expressing with the variable X as "eat X". The superiority of this approach is that variable X can accommodate a variety of substitute words such as apple and pineapple. Prereordered-based SMT is a word order change before translation. In the case of Korean, the word order of the sentence is Subject-Object-Verb (SOV), while English is Subject-Verb-Object (SVO). If the word order is different, this is a methodology to alter the word order in accordance with the word order of the target language to be translated before proceeding with the translation. Syntax Base SMT is a translation technique that changes from "eat X" to "eat NP (Noun Phrase)" in the Hierarchical Phase-Based SMT. In other words, not all phrases can come to the candidate group, but only nouns can be placed to the candidate group, and it eliminates unnecessary translation candidates in advance \cite{zens2002phrase,koehn2009statistical}.

~\\
\textbf{Neural Machine Translation}
NMT uses deep neural network to translation system. Based on the Sequence to Sequence model, the source language is vectorized through encoder and the latent vector is untangled through decoder to generate the target language. It is a method of utilizing deep neural network to uncover the most appropriate representations and translation results with a single pair of statements of input and output. For the text-to-text sequential modeling \cite{sutskever2014sequence}, NMT model generally comprises encoder and decoder structure that takes input sequence and generates output sequence auto-regressively. It has been developed to Recurrent Neural Network (RNN) \cite{cho2014learning,bahdanau2014neural}, Convolution Neural Network (CNN) \cite{gehring2017convolutional,wu2019pay}, and Transformer-based model \cite{vaswani2017attention} which outperforms other existing methods. Furthermore, fine-tuning approaches for pre-trained language models have recently shown the best performance including Cross-lingual Language Model Pre-training (XLM) \cite{lample2019cross}, Masked Sequence to Sequence Pre-training for Language Generation (MASS) \cite{song2019mass}, and Multilingual BART (mBART) \cite{liu2020multilingual}. In contrast, the parameters and model sizes of these pre-trained language models are extremely large for real-world industries to deploy the services. To address this issue, we present that the optimal model to proceed with the service is Transformer, considering the overall factors such as model performance, speed, and memory in recently published papers, and conduct experiments based on that model.

\begin{table*}[!tbh]
\renewcommand{\arraystretch}{1}
\centering
\caption{\label{tab:feature} Overview of features in LIWC}
\scalebox{0.8}{

\begin{tabular}{c|c}
\toprule
Category & Features(Label) \\ \midrule \hline
Summary  & Analytical thinking(Analytic), Clout(Clout),   \\
language variables & Authenticity(Authentic), Emotional tone(Tone) \\
\hline
  & Words per sentence(WPS), Percent of target words captured by the dictionary(Dic), \\
Linguistic& Percent of words in the text that are longer than six letters(Sixltr), Word count(WC),  \\
Dimension & Articles(article), Prepositions(prep), Total pronouns(pronoun), Personal pronouns(ppron), \\
 &   1st pers singular(i), 1st pers plural(we), 2nd person(you), \\
& 3rd pers singular(shehe), 3rd pers plural(they), Impersonal pronouns(ipron)  \\
\hline
  & Auxiliary verbs(auxverb), Common verbs(verb), Common Adverbs(adverb), \\
Grammars & Conjunctions(conj), Negations(negate), Common adjectives(adj), Comparisons(compare), \\
 & Interrogatives(interrog), Number(number), Quantifiers(quant)  \\ \hline
Affect process & Total affect process(affect), Positive emotion(posemo), \\
& Negative emotion(negemo), Anxiety(anx), Anger(anger), Sadness(sad)  \\ \hline
Cognitiive process & Total cognitiive process(cogproc), Insight(insight), Cause(cause),\\
& Discrepanices(discrep), Tentativeness(tentat), Certainty(certain), Differentiation(differ)  \\ \hline
Social process & Total social process(social), Familty(family), Friends(friend), \\
& Female referents(female), Male referents(male)  \\ \hline
Perceptual process & Total perceptual process(percept), Seeing(see), Hearing(hear), Feeling(feel)  \\ \hline
Biological process & Total biological process(bio), Body(body), Health/Illness(health),\\
& Sexuality(sexual), Ingesting(ingest)  \\ \hline
Drives & Total drives(drives), Affiliation(affiliation), Achievement(achieve), \\
& Power(power), Reward focus(reward), Risk focus(risk)  \\ \hline
Time orientations & Past focus(focuspast), Present focus(focuspresent), Future focus(focusfuture)  \\ \hline
Relativity & Total relativity(relativ), Motion(motion), Space(space), Time(time)  \\ \hline
Personal concerns & Work(work), Home(home),  Money(money), \\
& Leisure activities(leisure),Religion(relig), Death(death)  \\ \hline
Informal & Total informal language markers(Informal), Assents(assent),\\
language markers &  Fillers(filler), Swear words(swear), Netspeak(netspeak), Nonfluencies(nonfl)  \\ \hline
& Total punctuation(Allpunc), Semicolons(SemiC), Commas(Comma),  Colons(Colon),  \\
Punctuations & Parantheses(Parenth), Question marks(QMark), Exclamation marks(Exclam), \\
& Periods(Period), Apostrophes(Apostro), Quoatation marks(Quote)  \\
& Dashes(Dash), Other puntuation(OtherP)  \\ \hline
\bottomrule
\end{tabular}
}
\end{table*}

\subsection{AI Hub}
With the advent of the fourth industrial revolution\cite{schwab2017fourth}, inter-language exchange of information has rapidly been increased, accelerating the demand for the development of advanced translation systems. Despite the development of automatic translation systems accompanied by the growth of Information Technology (IT), there remain several difficulties in the industrial services of machine translation. The cost and time barriers of building early translation solutions exist and it is difficult to obtain quality data. Moreover, there are challenges in maintaining NMT performance quality, obstacles to obtaining domain-specific language pairs, and struggling to provide domain-specific NMT solutions. In other words, most of the difficulties are due to the lack of translation data, namely parallel corpus. Additionally, intellectual property makes it complicated to secure data, and there are numerous costs to collect, which is a major challenge for start-ups in the artificial intelligence-based industry or companies preparing for innovation \cite{park2021study}.

In general, a single corpus is relatively uncomplicated to obtain and sufficient amount can be secured, but in the case of parallel corpus, it becomes tough to acquire. Furthermore, constructing parallel corpus requires a number of high-level techniques for refining, pre-processing practical original corpus, and translating a single corpus into a desired heterogeneous language demands a lot of expenses. 

To mitigate these limitations, AI Hub constructs and continuously distributes public data nationally. AI Hub is a platform that integrates AI infrastructure such as AI data, AI software, algorithms, and computing resources that are essential for developing AI technologies, products, and services. It is also releasing data related to image recognition as well as data related to machine reading, machine translation, and voice recognition. This platform contributes to the creation of an intelligence information society and an artificial intelligence industrial ecosystem including medium-sized venture enterprises, research institutes, and individuals in Korea by disclosing high-quality and high-capacity artificial intelligence data.

AI Hub has released several datasets on MT, including the high-quality Korean-English corpus released in 2019 and 2021. Subsequently, the construction of parallel corpus, including Korean-Japanese. Korean-Chinese, and other Korean-language parallel corpus has been actively established. However, close verification of these data is not being done specifically, and we seeks to proceed with quality confirmation by building a LIWC and a real-world NMT model.

\subsection{Parallel Corpus Quality Assessment}
Accompanied by the increase of publicly released parallel corpus such as FLORES-101\cite{goyal2021flores} and AI hub, the importance of evaluating and improving the quality of the parallel corpus becomes higher. Especially for the data construction process, assessing the quality of the corpus is regarded as an essential process. For example, in the case of AI hub, all publicly available corpora were constructed through a multi-phase process that proceeded with machine translation followed by the human examination. For the corresponding examination process, semantic coherence and sentence alignment are mainly inspected. This can be viewed as checking its suitability to the intended purpose of the data construction.

However, as data acquisition becomes more accessible and the amount of data used for training increases, exploring each data with human labor leads to considerable cost. For instance, as the total amount of parallel corpus released by the AI hub is approximately 7M, it can be expected that tremendous time and cost are required to examine the whole corpus.

For the alleviation of such limitation, corpus evaluation studies were being made by mainly focusing on the minimization of direct human examination. Representative methods include the use of several translation rules established in advance \cite{espla2019paracrawl}, the gale church algorithm \cite{gale1993program} that evaluates the overall align of the sentence, and the Bilingual Sentence Aligner \cite{simard1998bilingual}. The validity of these corpus evaluation methodologies is generally evaluated based on the performance of the MT system generated through the corresponding corpus. In particular, evaluation criteria, such as sentence alignment, were confirmed to be effective as corpus evaluation metric through the performance verification of the SMT model generated through the corpus \cite{abdul2012extrinsic}. Furthermore, with the development of deep neural modeling, these methodologies are evaluated by the performance of NMT model \cite{espla2019paracrawl}.

However, most of these studies aim to improve the performance of the MT system itself trained by the corresponding parallel corpus. These often led to inconsistent results where several data that was not considered to be noisy in training SMT system considerably deteriorated the performance of the NMT system when utilized in training process \cite{khayrallah2018impact}. Thus, it shows that these corpus evaluation criteria may not be consistent enough to be directly related to actual quality assessment. In this study, we analyze the corpus using a sentence analysis tool called LIWC, which has not been utilized as a parallel corpus inspection and as an objective evaluation index for the corpus quality. Also, following previous studies, we check the performance of the NMT system trained through the parallel corpus and analyze the characteristics and quality of the corresponding corpus that can be obtained through the result.

\subsection{Korean Neural Machine Translation Research}
Recently, various services are being provided in South Korea as well as MT-related research. Along with Papago translation service \cite{lee2016papago} which is serviced by Naver corporation, MT services are conducted by many companies and laboratories, including the Electronics and Telecommunications Research Institute (ETRI), Kakao, SYSTRAN and Genie Talk at Hancom Interfree. 

Research on NMT data pre-processing is mainly being conducted in the academia of Korea University. There are several related studies including Onepiece which proposes a specialized sub-word tokenization in Korean \cite{park2021should} and applying PCF to the Korean-English NMT for the first time \cite{pcj01}. They also propose a methodology for training with relative ratio when configuring batch rather than simply applying back translation or copied translation when applying data augmentation \cite{pcj02}. This results in higher performance than simply using back translation. In addition, based on machine translation, they have conducted various applications such as Korean spelling corrector \cite{park2020neural}, English grammar corrector \cite{park2020ancient}, and cross lingual transfer learning \cite{lee2021exploring}. In conclusion, there are various experiments and studies based on the importance of pre-processing and data augmentation as well as research on NMT models.

\section{Analyzing the AI Hub Corpus Using LIWC}
\label{sec:LIWC}
\subsection{Linguistic Inquiry and Word Count (LIWC)} 
LIWC is a natural-language analysis software, which allows for the investigation of various emotional, cognitive, and structural components of specific sentences \cite{pennebaker2015development}. LIWC offers corpus analysis by referring to a dictionary comprising 93 features. Each feature provided is shown in Table \ref{tab:feature}. Every feature can be classified into 14 categories: summary language variables, linguistic dimensions, grammar, affect process, cognitive process, social process, perceptual process, biological process, drives, time-orientations, relativity, personal concerns, informal language markers, and punctuations. This is different from the classifications presented in the LIWC manual, which are classified into 16 categories for more intuitive analysis. We consolidate them into new categories to avoid confusion and achieve our objective. We merged auxiliary verbs, common adverbs, conjunctions, and negations of function words in grammar with Other Grammar such as conjunctions, adjectives, and so on defined in the initial categories of Manual. Pronouns, articles, and prepositions, which involve functional words, were joined with the linguistic dimension, thereby helping in the understanding of text through the rules of sentence structure. Grammar represents the grammatical components of a sentence and comprises some parts of speech. The summary language variable represents the summarized value of all the linguistic features representing the overall features of a sentence. The affect process quantifies emotions and feelings. The biological process category represents biological topics, such as body, health, and ingest in text. The drives category represents motivations and needs, which appear in text. The time-orientation category helps in the understanding of the tense used in text because LIWC contains both the tenses of verbs and general time orientations. The relativity category represents relatively-trivial topics and personal concerns as well as literal meanings of the concerned topics in text. The punctuations and informal language markers have similar meanings. In this study, we conducted an in-depth analysis with respect to the following five aspects. 

First, morphological analysis can be conducted by referring to morphological features, such as grammar and linguistic dimensions. Second, the investigation of summary language variables, general descriptors, time orientation, punctuation, and informal language markers categories, enables the analysis of sentence syntax. Third, semantic analysis through the inspection of various topics, including cognitive, social, perceptual, and biological process, as well as relativity and personal concerns, can be implemented. Fourth, we can conduct sentimental analysis through the affect process, which involves positive and negative emotions. Finally, the social category, which contains male and female referents, enables the analysis of gender bias. Specifically, in the field of NMT, numerous studies have been conducted to reduce the prevalence of gender bias \cite{prates2018assessing, saunders2020reducing}. In the future, this approach can be used to inspect the performance of MT systems.

LIWC is mainly leveraged in the field of psychology, especially during the investigation of linguistic characteristics revealed in the writings of psychiatric patients \cite{tausczik2010psychological, coppersmith2014quantifying}. Furthermore, LIWC has been recently utilized in numerous studies on natural language processing (NLP), and its effectiveness and relevance in the field of NLP has been demonstrated. For instance, the performance of misinformation detection \cite{su2020motivations}, sentiment analysis, and plagiarism detection \cite{garcia2020using} can be improved by applying LIWC, and the effectiveness of LIWC can be evaluated through its comparison to BERT \cite{biggiogera2021bert}. Following these trends, we aim to investigate all the parallel corpora for Korean released by AI Hub through morphological, semantic, sentence-syntactic, sentimental, and gender-bias aspects.

\subsection{Korean-English Parallel Corpus}

\label{sec:ko-en-para}
\begin{figure}[h!]
  \centering
  \includegraphics[scale=0.60]{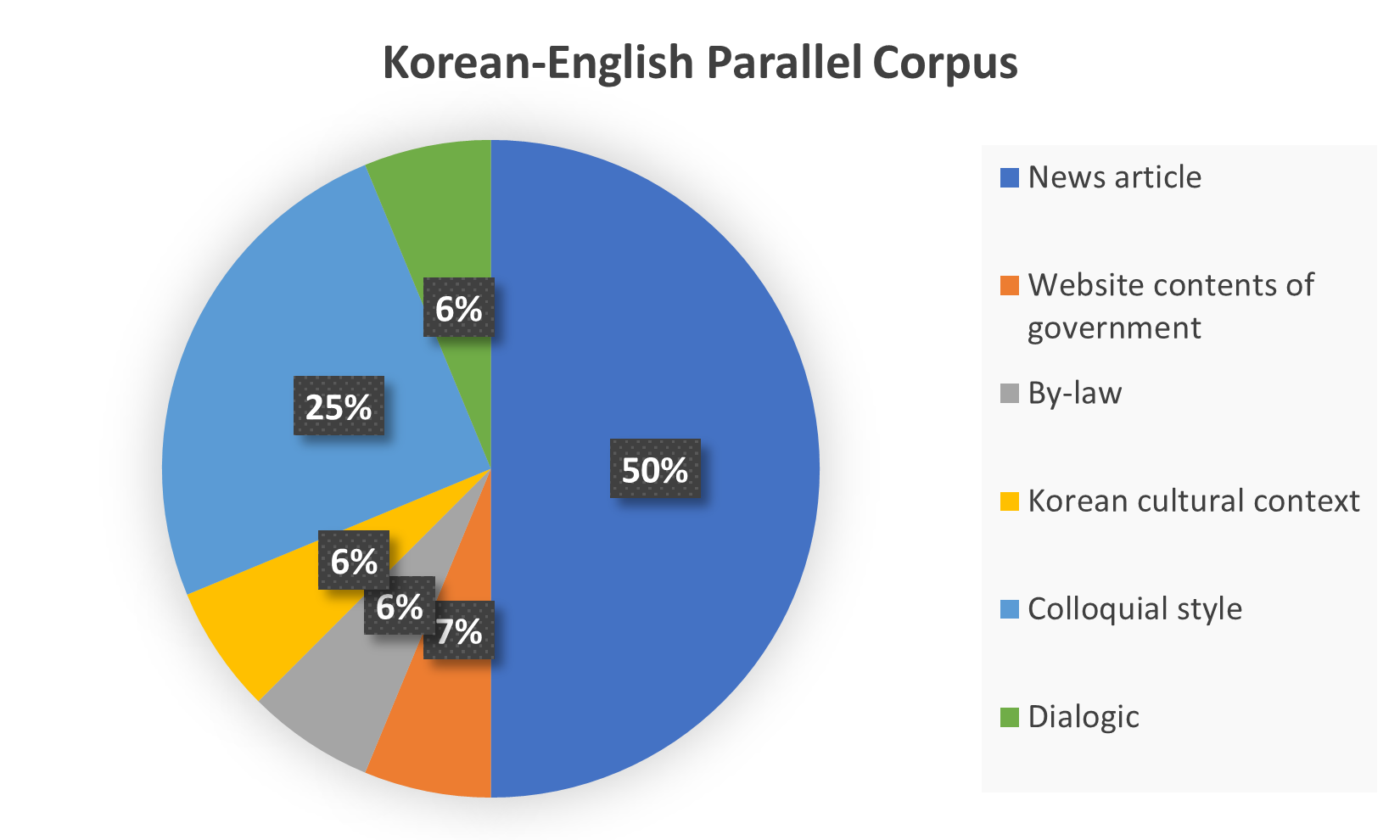}
  \caption{Data-domain statistics of the Korean-English Parallel Corpus.}
  \label{fig:mt}
\end{figure}

~\\
\paragraph{Corpus Description}
The Korean-English parallel corpus \footnote{\url{https://aihub.or.kr/aidata/87}} is a parallel corpus from AI Hub, which was released in 2019. The Corresponding corpus was built through the cooperation of Saltlux partners \footnote{\url{http://saltlux.com}}, Flitto \footnote{\url{https://www.flitto.com}}, and Evertran \footnote{\url{http://www.evertran.com}}. The total amount of sentence pairs in the constructed corpus is 1.6M, which comprises 800K news articles, 100K website contents from the government, 100K instances of by-law data, 100K instances of Korean-based cultural contexts, 400K instances of colloquial-style data, and 100K instances of dialogic data. The ratios of each domain to the entire corpus are shown in Figure \ref{fig:mt}. It can be considered the most representative Korean-English parallel corpus, and many Korean-related studies on MT have been conducted based on that corpus \cite{moon2021filter}. 

For a more thorough data analysis, we conducted an in-depth investigation of this corpus with respect to various features, such as morphemes, syntax information, and the characteristics of the corpus. We analyzed such features using LIWC, and the results are shown in Table \ref{tab:3.2}.

\begin{table*}[!tbh]
\renewcommand{\arraystretch}{0.6}
\centering
\caption{\label{tab:3.2} LIWC results of Korean-English Domain-Specialized Parallel and Korean-English Parallel Corpus. }
\scalebox{0.7}{
\begin{tabular}{l|l|l|l|l|l}
\hline
                                        &              & \multicolumn{3}{l|}{Ko-En Domain-Specialized Parallel Corpus} & Ko-En Parallel Corpus \\ \hline \hline
Categories                              & Features & train                & valid               & total               & total                             \\ \hline 
\multirow{4}{*}{summary language variables}      & \textbf{Analytic}     & \textbf{97.14}            &  \textbf{97.26}               & \textbf{97.16}               & \textbf{94.16}                             \\ \cline{2-6} 
                                        & Clout        & 61.1                 & 61.17               & 61.11               & 65.39                             \\ \cline{2-6} 
                                        & Authentic    & 25.7                 & 25.37               & 25.66               & 27.76                             \\ \cline{2-6} 
                                        & Tone         & 49.14                & 49.08               & 49.13               & 54.48                             \\ \hline
\multirow{15}{*}{Linguistic dimensions} & WC           & 33,750,816           & 4,222,191           & 37,973,007          & 38,481,936                        \\ \cline{2-6} 
                                        & WPS          & 27.33                & 27.29               & 27.33               & 25.39                             \\ \cline{2-6} 
                                        & Sixltr       & 26.5                 & 26.49               & 26.5                & 25.69                             \\ \cline{2-6} 
                                        & Dic          & 76.33                & 76.3                & 76.32               & 79.25                             \\ \cline{2-6} 
                                        & function     & 43.78                & 43.78               & 43.78               & 45.24                             \\ \cline{2-6} 
                                        & pronoun      & 4.9                  & 4.87                & 4.89                & 6.73                              \\ \cline{2-6} 
                                        & ppron        & 1.54                 & 1.54                & 1.54                & 3.08                              \\ \cline{2-6} 
                                        & \textbf{i}            & \textbf{0.21}                 & \textbf{0.21}                & \textbf{0.21}                & \textbf{1.02}                              \\ \cline{2-6} 
                                        & we           & 0.23                 & 0.23                & 0.23                & 0.38                              \\ \cline{2-6} 
                                        & you          & 0.3                  & 0.3                 & 0.3                 & 0.56                              \\ \cline{2-6} 
                                        & shehe        & 0.42                 & 0.42                & 0.42                & 0.67                              \\ \cline{2-6} 
                                        & they         & 0.38                 & 0.38                & 0.38                & 0.45                              \\ \cline{2-6} 
                                        & ipron        & 3.36                 & 3.33                & 3.35                & 3.64                              \\ \cline{2-6} 
                                        & article      & 10.4                 & 10.42               & 10.4                & 10.2                              \\ \cline{2-6} 
                                        & prep         & 15.51                & 15.52               & 15.51               & 15.1                              \\ \hline
\multirow{10}{*}{grammar}               & auxverb      & 6.07                 & 6.05                & 6.06                & 6.75                              \\ \cline{2-6} 
                                        & adverb       & 2.46                 & 2.47                & 2.46                & 2.55                              \\ \cline{2-6} 
                                        & conj         & 5.92                 & 5.92                & 5.92                & 5.43                              \\ \cline{2-6} 
                                        & negate       & 0.63                 & 0.62                & 0.63                & 0.7                               \\ \cline{2-6} 
                                        & \textbf{verb}         & \textbf{9.94}                 & \textbf{9.91}                & \textbf{9.94}                & \textbf{11.36}                             \\ \cline{2-6} 
                                        & adj          & 4.49                 & 4.46                & 4.49                & 4.47                              \\ \cline{2-6} 
                                        & compare      & 2.35                 & 2.34                & 2.35                & 2.25                              \\ \cline{2-6} 
                                        & interrog     & 1.21                 & 1.2                 & 1.21                & 1.3                               \\ \cline{2-6} 
                                        & number       & 3.27                 & 3.28                & 3.27                & 2.53                              \\ \cline{2-6} 
                                        & quant        & 1.36                 & 1.36                & 1.36                & 1.4                               \\ \hline
\multirow{6}{*}{affective process}      & affect       & 3.78                 & 3.75               & 3.77                & 4.03                             \\ \cline{2-6} 
                                        & \textbf{posemo}       & \textbf{2.49}                 & \textbf{2.48}                & \textbf{2.49}                & \textbf{2.75}                              \\ \cline{2-6} 
                                        & negemo       & 1.24                 & 1.23                & 1.24                & 1.23                              \\ \cline{2-6} 
                                        & anx          & 0.17                 & 0.17                & 0.17                & 0.19                              \\ \cline{2-6} 
                                        & anger        & 0.21                 & 0.2                 & 0.21                & 0.29                              \\ \cline{2-6} 
                                        & sad          & 0.3                  & 0.3                 & 0.3                 & 0.28                              \\ \hline
\multirow{5}{*}{social process}         & social       & 5.07                 & 5.04                & 5.06                & 6.7                               \\ \cline{2-6} 
                                        & family       & 0.23                 & 0.23                & 0.23                & 0.25                              \\ \cline{2-6} 
                                        & friend       & 0.12                 & 0.12                & 0.12                & 0.15                              \\ \cline{2-6} 
                                        & female       & 0.18                 & 0.17                & 0.18                & 0.34                              \\ \cline{2-6} 
                                        & \textbf{male}         & \textbf{0.46}                 & \textbf{0.47}                & \textbf{0.46}                & \textbf{0.63}                              \\ \hline
\multirow{7}{*}{cognitive process}      & \textbf{cogproc}      & \textbf{6.75}                 & \textbf{6.73}                & \textbf{6.74}                & \textbf{7.48}                              \\ \cline{2-6} 
                                        & insight      & 1.56                 & 1.56                & 1.56                & 1.75                              \\ \cline{2-6} 
                                        & cause        & 1.69                 & 1.68                & 1.69                & 1.73                              \\ \cline{2-6} 
                                        & discrep      & 0.77                 & 0.77                & 0.77                & 0.99                              \\ \cline{2-6} 
                                        & tentat       & 1.14                 & 1.14                & 1.14                & 1.41                              \\ \cline{2-6} 
                                        & certain      & 0.65                 & 0.65                & 0.65                & 0.7                               \\ \cline{2-6} 
                                        & differ       & 1.92                 & 1.91                & 1.92                & 1.96                              \\ \hline
\multirow{4}{*}{perceptual process}     & \textbf{percept}      & \textbf{1.53}                 & \textbf{1.53}                & \textbf{1.53}                & \textbf{1.82}                              \\ \cline{2-6} 
                                        & see          & 0.58                 & 0.58                & 0.58                & 0.68                              \\ \cline{2-6} 
                                        & hear         & 0.48                 & 0.48                & 0.48                & 0.65                              \\ \cline{2-6} 
                                        & feel         & 0.32                 & 0.32                & 0.32                & 0.35                              \\ \hline
\multirow{5}{*}{Biological process}     & \textbf{bio}          & \textbf{2.31}                 & \textbf{2.31}                & \textbf{2.31}                & \textbf{1.63}                              \\ \cline{2-6} 
                                        & body         & 0.46                 & 0.46                & 0.46                & 0.46                              \\ \cline{2-6} 
                                        & health       & 1.35                 & 1.36                & 1.35                & 0.67                              \\ \cline{2-6} 
                                        & sexual       & 0.04                 & 0.04                & 0.04                & 0.05                              \\ \cline{2-6} 
                                        & ingest       & 0.51                 & 0.51                & 0.51                & 0.46                              \\ \hline
\multirow{6}{*}{drives}                 & \textbf{drives}       & \textbf{7.39}                 & \textbf{7.4}                 & \textbf{7.4}                 & \textbf{8.26}                              \\ \cline{2-6} 
                                        & affiliation  & 1.52                 & 1.52                & 1.52                & 1.77                              \\ \cline{2-6} 
                                        & achieve      & 1.92                 & 1.92                & 1.92                & 1.96                              \\ \cline{2-6} 
                                        & power        & 3.33                 & 3.33                & 3.33                & 3.94                              \\ \cline{2-6} 
                                        & reward       & 1.02                 & 1.02                & 1.02                & 1.05                              \\ \cline{2-6} 
                                        & risk         & 0.79                 & 0.78                & 0.79                & 0.63                              \\ \hline
\multirow{3}{*}{time-orientations}      & focuspast    & 3.27                 & 3.25                & 3.26                & 3.4                               \\ \cline{2-6} 
                                        & focuspresent & 5.79                 & 5.78                & 5.79                & 6.63                              \\ \cline{2-6} 
                                        & focusfuture  & 0.97                 & 0.96                & 0.96                & 1.49                              \\ \hline
\multirow{4}{*}{relativivity}           & relativ      & 14.53                & 14.55               & 14.53               & 14.29                             \\ \cline{2-6} 
                                        & motion       & 1.72                 & 1.72                & 1.72                & 1.81                              \\ \cline{2-6} 
                                        & \textbf{space}        & \textbf{8.31}                 & \textbf{8.32}                & \textbf{8.31}                & \textbf{7.83}                              \\ \cline{2-6} 
                                        & time         & 4.59                 & 4.6                 & 4.59                & 4.73                              \\ \hline
\multirow{6}{*}{personal concerns}      & \textbf{work}         & \textbf{5.64}                 & \textbf{5.64}                & \textbf{5.64}                & \textbf{6.37}                              \\ \cline{2-6} 
                                        & leisure      & 1.52                 & 1.52                & 1.52                & 1.43                              \\ \cline{2-6} 
                                        & home         & 0.52                 & 0.53                & 0.52                & 0.5                               \\ \cline{2-6} 
                                        & money        & 2.19                 & 2.18                & 2.19                & 1.87                              \\ \cline{2-6} 
                                        & relig        & 0.23                 & 0.23                & 0.23                & 0.29                              \\ \cline{2-6} 
                                        & death        & 0.14                 & 0.14                & 0.14                & 0.14                              \\ \hline
\multirow{6}{*}{informal language}      & informal     & 0.23                 & 0.23                & 0.23                & 0.26                              \\ \cline{2-6} 
                                        & swear        & 0                    & 0                   & 0                   & 0.01                              \\ \cline{2-6} 
                                        & netspeak     & 0.1                  & 0.11                & 0.1                 & 0.1                               \\ \cline{2-6} 
                                        & assent       & 0.06                 & 0.06                & 0.06                & 0.09                              \\ \cline{2-6} 
                                        & nonflu       & 0.08                 & 0.08                & 0.08                & 0.09                              \\ \cline{2-6} 
                                        & \textbf{filler}       & \textbf{0}                    & \textbf{0}                   & \textbf{0}                   & \textbf{0}                                 \\ \hline
\multirow{12}{*}{punctuations}           & AllPunc      & 14.72                & 14.72               & 14.72               & 14.68                             \\ \cline{2-6} 
                                        & Period       & 3.81                 & 3.81                & 3.81                & 4.33                              \\ \cline{2-6} 
                                        & Comma        & 6.23                 & 6.23                & 6.23                & 5.24                              \\ \cline{2-6} 
                                        & \textbf{Colon}        & \textbf{0.03}                 & \textbf{0.03}                & \textbf{0.03}                & \textbf{0.08}                              \\ \cline{2-6} 
                                        & SemiC        & 0.01                 & 0.01                & 0.01                & 0.01                              \\ \cline{2-6} 
                                        & QMark        & 0.01                 & 0.01                & 0.01                & 0.22                              \\ \cline{2-6} 
                                        & Exclam       & 0                    & 0                   & 0                   & 0                                 \\ \cline{2-6} 
                                        & Dash         & 1.88                 & 1.89                & 1.89                & 1.6                               \\ \cline{2-6} 
                                        & Quote        & 1.11                 & 1.1                 & 1.11                & 1.16                              \\ \cline{2-6} 
                                        & Apostro      & 0.85                 & 0.85                & 0.85                & 1.11                              \\ \cline{2-6} 
                                        & Parenth      & 0.56                 & 0.57                & 0.56                & 0.7                               \\ \cline{2-6} 
                                        & OtherP       & 0.22                 & 0.23                & 0.22                & 0.23                              \\ \hline
\end{tabular}}
\end{table*}

\paragraph{Corpus Analysis}
For the morphological analysis, we conducted morphological analysis through the linguistic dimension and the grammar of linguistic-feature results obtained using LIWC. In the linguistic dimension category, it shows high frequency in `prepositions (prep)' and `articles (article)' of the total part-of-speech, at 15.1\% and 10.2\%, respectively. Additionally, the `auxiliary verb (auxverb)' in the grammar category appears for 6.75\%, and its prevalence is higher than that of others, such as `commonverb (verb),' which shows the highest frequency, at 11.36\%. 
This result indicates the continuous prevalence of be verbs (am, are, is, was, and were), and the perfect tense of English characteristics, such as diverse tenses and conjugations, was reflected as data, rather than a base form of a verb.

In personal-pronoun analysis, which consists of \{`1st pers singular (i)', `1st pers plural (we)', `2nd person (you)', `3rd pers singular (shehe)', `3rd pers plural (they)', and `impersonal pronouns (ipron)'\}, the frequency of impersonal pronouns (ipron) is similar to that of `personal pronouns (ppron)', and the `1st pers singular(i)' has the highest frequency as one of the personal pronouns. 
The `2nd person' and `3rd pers singular' pronouns also came next in order of prevalence. Unlike other corpora in which impersonal pronouns are predominant, this corpus includes both colloquial and dialogic sentences because it comprises interactive conversations between the first-person perspective and other person perspectives.

Syntactic analysis is defined as an analytic approach that informs us on the grammatical meaning of specific sentences or parts of such sentences. This approach avails the type of tone and atmosphere used in sentences through the summary language variables category. We also obtain the sentence length from `word count (wc)', 'word per sentence (wps)', and lengthy word count using `Sixltr'. We explore the sentences based on whether they are represented using statements, questions, or quotations using the punctuation category. We use the time orientations category to understand the point of view, and we investigate the 'assents (assent)', `fillers (filler)', `swear words (swear)' in the informal language markers category. In the end, these features contribute to understanding the syntactic information of the corpus.

We show the `thinking (analytic)', `clout (clout)' (i.e., the representation of trust), `authenticity (authentic)' (i.e., the representation of sincerity), and `emotional tone (tone)' as 94.16\%, 65.39\%, 27.76\%, and 54.48\%, respectively. We can find that the prevalence of `analytic' is relatively low whereas that of `clout', `authentic', and `tone' is relatively high. This reflects the characteristics of each component in the corpus that contains various descriptive styles (\emph{i.e.} colloquial or literary), rather than focusing on conveying and explaining specific domain knowledge.
Various descriptive styles of this corpus can also be found by inspecting punctuation category which shows relatively high appearance rate of `question marks (QMark)'.

In the results of the linguistic dimensions category, we establish that the average word count of a sentence is 25.39, and `Sixltr' accounts for 25\% of the total word count. The articles and prep categories also account for a large proportion at 10.4\% and 15.5\%, respectively, and as a result of the analysis of the time orientations category, the ratios are high in the order of present focus, past focus, and future focus. Additionally, in the grammar category, words that directly represent `number (number)' are used approximately twice as often as `quantifiers (quant)' representing quantities.

In the informal language mark category, the values are higher than those of the other categories. It is noteworthy that `swear words', such as `damn' and `shit,' and `fillers,' such as `you know' and `i mean,' which are used as interludes between conversations, are close to zero in other corpora results. It seems that this is because the corpus contains written, colloquial, and dialogue language, unlike general data, which consists only of written or spoken language.

In semantic analysis, semantics is the study of analyzing meaning in units of text, sentences, and phrases. In this study, we attempt to understand the corpus in depth by checking which of the various topics, such as drives and biological process, has a high ratio.

In the corpus, the relativity category corresponding to a relatively trivial topic was found to be at the highest level at 14.29\%, and `space (space)' accounted for the highest prevalence at 7.83\%. Next are the drives, cognitive process, personal concerns, and social process categories in that order. Specifically, in the personal concerns category, `work (work)' occupied more than half. This result is because, unlike other corpora, this corpus includes colloquial words and dialogues. For this reason, the relatively trivial topic, which is the topic of conversation, and the individual's sense of purpose, thoughts, and interests are relatively clearly revealed.

Sentiment analysis involves analyzing the degree of positivity and negativity appearing in text. LIWC supports the analysis of the scale of `positive emotion (posemo)' and `negative emotion (negemo), such as `anger,' `anxiety,' and `sadness.'

`posemo' occurs twice as much as `negemo' in this corpus. Specifically, as a whole, words expressing emotions were used the most compared to other data, thereby revealing the characteristics of the corpus, which includes dialogues and spoken words.

Gender bias is an important factor when it comes to determining the quality of MT. The results of `Female referents (female)' and `Male referents (male)' in the social process category represent the feature of the level of being referent in the text. `Male' appears at a frequency of 0.63\%, which is twice that of `female' at 0.34\%. Although the gender balance of the corpus was not effectively achieved, it had the same results as the number of male and female referents presented through the LIWC average analysis results of various corpora, such as blogs, novels, and Twitter in the LIWC 2015 manual.

\subsection{Korean-English Domain-Specialized Parallel Corpus}
\label{sec:ko-en-domain}

\paragraph{Corpus Description}
The Korean-English Domain-Specialized Parallel Corpus \footnote{\url{https://aihub.or.kr/aidata/7974}} provides various parallel corpora specializing in several domains. This corpus was released in 2021, and three companies cooperated in its construction: Saltlux partners, Flitto, and Evertran.

\begin{figure}[h!]
  \centering
  \includegraphics[scale=0.60]{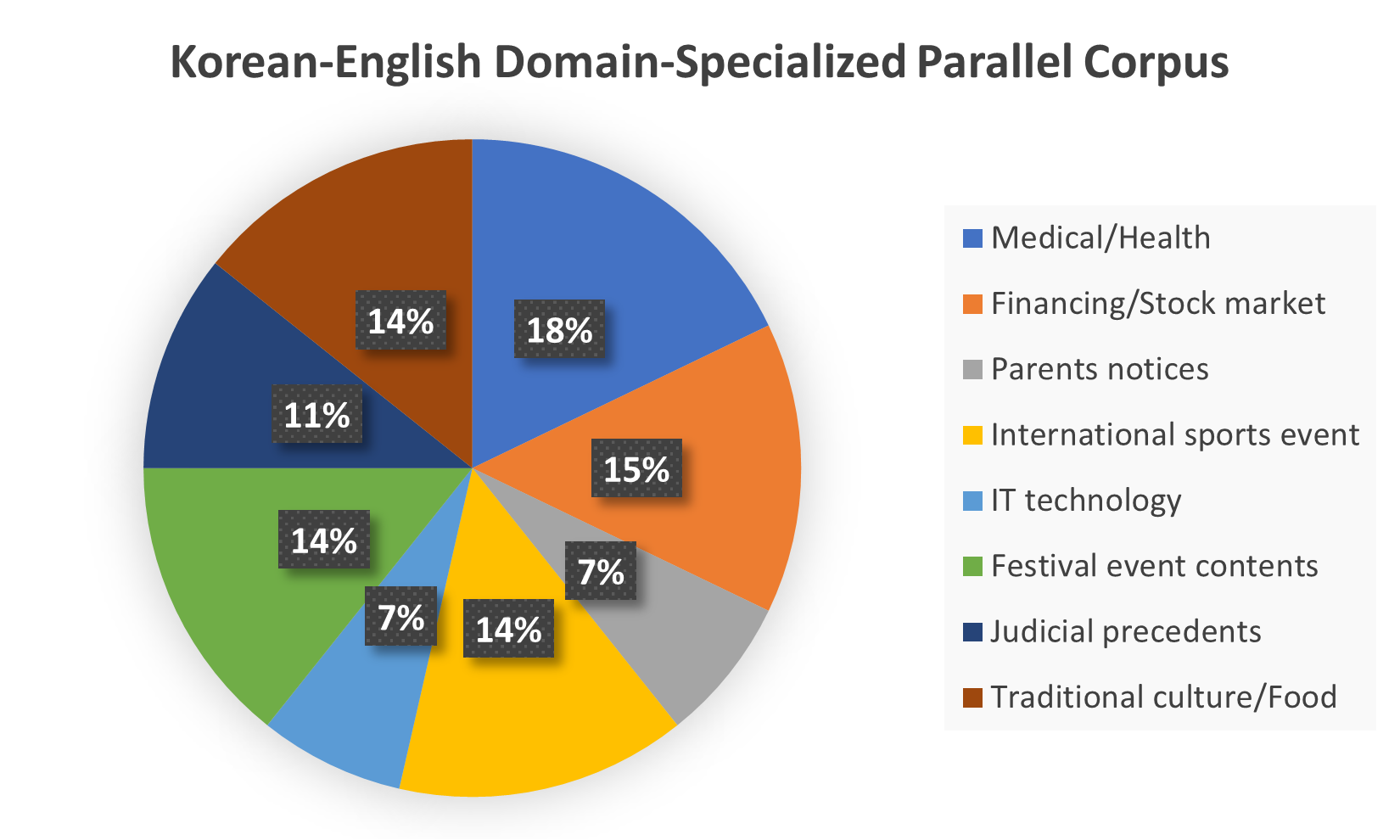}
  \caption{Data-domain statistics of the Korean-English Domain-Specialized Parallel Corpus.}
  \label{fig:domain}
\end{figure}

The corresponding corpus consists of 1.5M sentence pairs, including 250K instances of medical/health data, 200K instances of financial/stock market data, 100K instances of parent-notices data, 200K instances of international sport events data, and 100K instances of IT technology data, 200K instances of festival event content data, 150K judicial precedents, and 200K instances of data on traditional culture/food. The percentage of the domain data within the entire corpus is shown in Figure \ref{fig:domain}.

\paragraph{Corpus Analysis}
This corpus was released separately into training and validation datasets, and Table \ref{tab:3.2} shows the LIWC results of each dataset. The differences of the linguistic features between the training and validation datasets are generally less than 0.1\%. Although there are exceptional differences in some summary-language variables, such as `clout,' which means confidence, and `authentic,' which means authenticity, there are rarely any differences in these features because the differences are at approximately 1\%. Therefore we can conclude that the training and validation datasets are released in balance.

In morphological analysis, there are generally similar results to those of Section \ref{sec:ko-en-para}. However, impersonal pronouns are used twice as much as personal pronouns. Although all the results show that the use of personal pronouns is low, "i" and "you" show the lowest results in other corpora. 

Focusing on the syntactic category, `analytic', which represents analytic thinking in text, is highest in four summary language variables. The length of the sentences is the longest, and `commas' is the most frequently used in all the Korean-English corpora we analyzed. This is because there are many long sentences with several phrases explaining the domain-specialized concepts in such corpora. Time orientation is still the highest in the present and the lowest in the future. However, the difference between the present and the past tense is smallest in all corpora because the data contain past cases, such as judicial precedents and financial/stock market data.

Additionally, in semantics, the biological process category has the highest prevalence in the entire corpus. In this category, `health/illness (health)' is especially high because this corpus contains several domains that include both medical and international sport data. The results of `money (money)' and `leisure activities (leisure)' in the personal concerns category prove that there are international sports data and financial/stock market data in this corpus.

In the results of the sentimental analysis, `posemo' appears twice as much as `negamo'. Additionally, the use of words representing sentiments is 7\% lower than that presented in Section \ref{sec:ko-en-para}. Finally, male referents and female referents of this corpus allow for noticing gender bias because male referents are two times more than female referents unlike other corpora.

\subsection{Korean-English Parallel Corpus (Technology)}
\label{sec:ko-en-tech}

\paragraph{Corpus Description}
AI Hub released the technology-science domain-specialized Korean-English translation corpus \footnote{\url{https://aihub.or.kr/aidata/30719}} in 2021 through the cooperation of Twigfarm \footnote{\url{https://twigfarm.net/}}, Lexcode \footnote{\url{https://lexcode.co.kr/}}, Naver \footnote{\url{https://www.naver.com}}, the Korean telecommunications technology association (TTA) \footnote{\url{https://www.tta.or.kr/}}, and the Fun \& Joy company (FNJ) \footnote{\url{http://www.fnj.or.kr/home/index.html}}. The corresponding corpus was constructed for the support of ICT companies with respect to the translation of technical documents or product localization.

\begin{figure}[h!]
  \centering
  \includegraphics[scale=0.60]{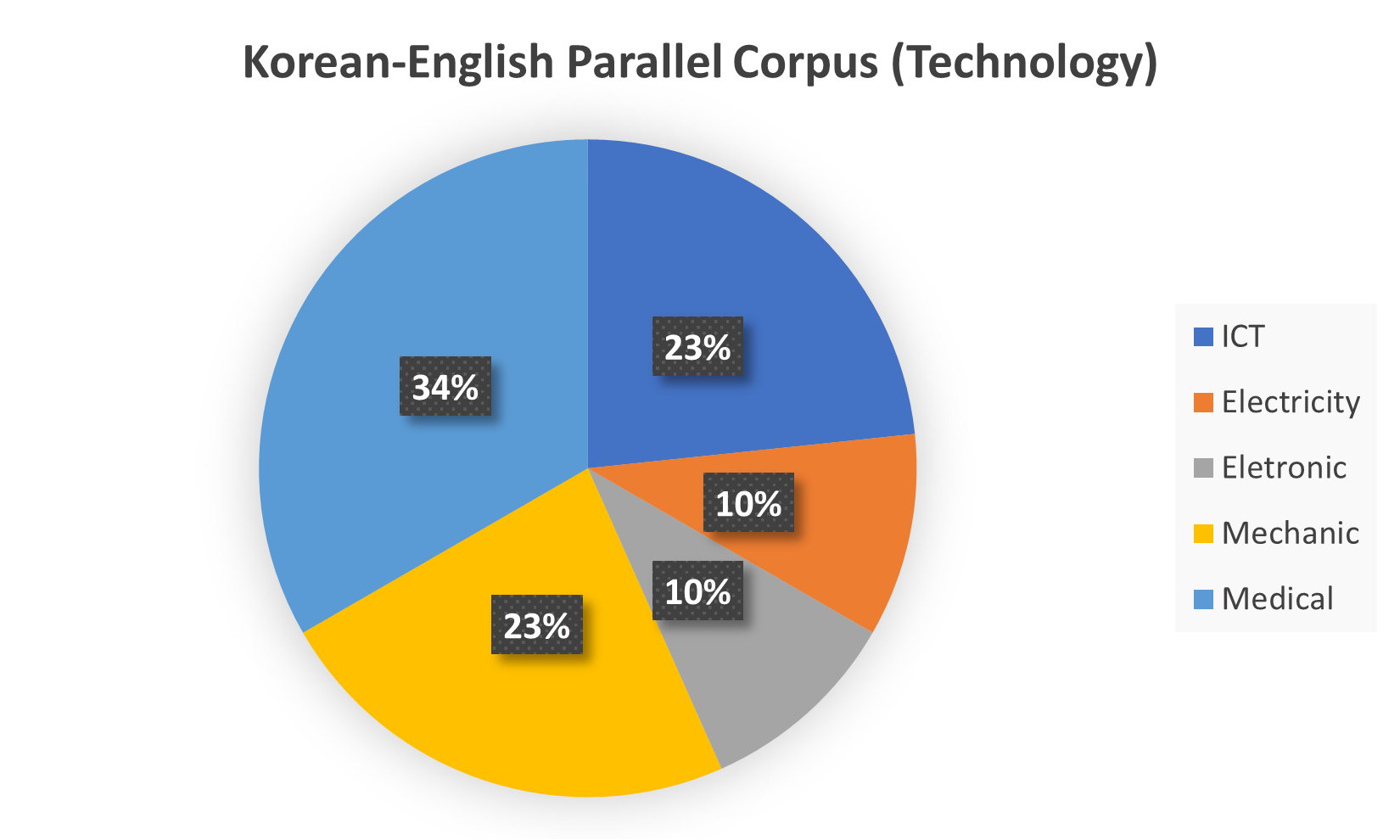}
  \caption{Data-domain statistics of the Korean-English Parallel Corpus (Technology).}
  \label{fig:data_ko_en_tech}
\end{figure}

The number of sentence pairs in the entire corpus is 1.5M, which comprises five domains, as shown in Figure \ref{fig:data_ko_en_tech}: 350K instances of ICT domain data, 150K instances of electricity domain data, 150K instances of electronic domain data, 350K instances of mechanical domain data, and 500K instances of medical domain data. For the construction of the high-quality corpus, expert-level revision by ICT professionals and several professors in translation fields was conducted after the initial corpus was compiled using a computer system. In this study, we partially leveraged 788K sentence pairs (ICT (35.2\%), mechanic (31.2\%), electricity (13.8\%), electronic (11.2\%), and medical (8.6\%)) because all the data is yet to be released.

\paragraph{Corpus Analysis}
The entire corpus comprises training and validation datasets, and the LIWC results are shown in Table \ref{tab:3.4}. 

The LIWC results show that there exist small differences between the training and validation datasets in terms of each feature value. We can also establish that the corresponding corpus contains more adverbs than other parallel corpora, and few personal pronouns have been used to the extent that impersonal pronouns (`ipron') have been used approximately 10 times more than personal pronouns. This allows us to identify the characteristics of the corpus that describe technology and phenomena compared to the corpora of other fields that explore people and culture.

Because the `Sixltr' rate is relatively high, whereas `WPS' is low, we can infer that short sentences, each of which consist long words, are mainly contained in the corpus. Furthermore, the corpus shows low `authentic' and emotional tone (`tone') rates. This indicates that by considering the characteristics of the technology domain, the representations of each sentence are concise. The present tense appears more frequently than the past or future tenses, thereby supporting the attributes of the technology domain that are mainly targeted to describe current technology, future complementary points, and expectations.

Compared to other corpora, the prevalence of the drive category, which represents the motivation of a sentence, is relatively low, whereas the biological processes category remains the most frequent. These results are contrary to those of the social domain corpus, which reflects the characteristics of the corresponding corpus.

The ratio of the conventional process to that of the perceptual process category was also the highest throughout all the corpora. Therefore, it can be confirmed that the characteristics of articles in the technology domain are presented in a manner that describes perceptions and cognitive processes, such as `insight,' `causation,' and `certainty,' about technology. One notable characteristic is the absence of gender bias. The ratio of the domains can vary because all the 1.5M sentences are completely constructed. Therefore, we can obtain the linguistic features of each domain by comparing the entire corpus to the present corpus in latter experiments.

\subsection{Korean-English Parallel Corpus (Social Science)}
\label{sec:ko-en-social}

\paragraph{Corpus Description}
Similar to the technology-science specialized corpus, the social science specialized Korean-English parallel corpus\footnote{\url{https://aihub.or.kr/aidata/30720}} was also published in 2021 through the cooperation of Twigfarm, Lexcode, Naver, TTA, and FNJ.

\begin{figure}[h!]
  \centering
  \includegraphics[scale=0.60]{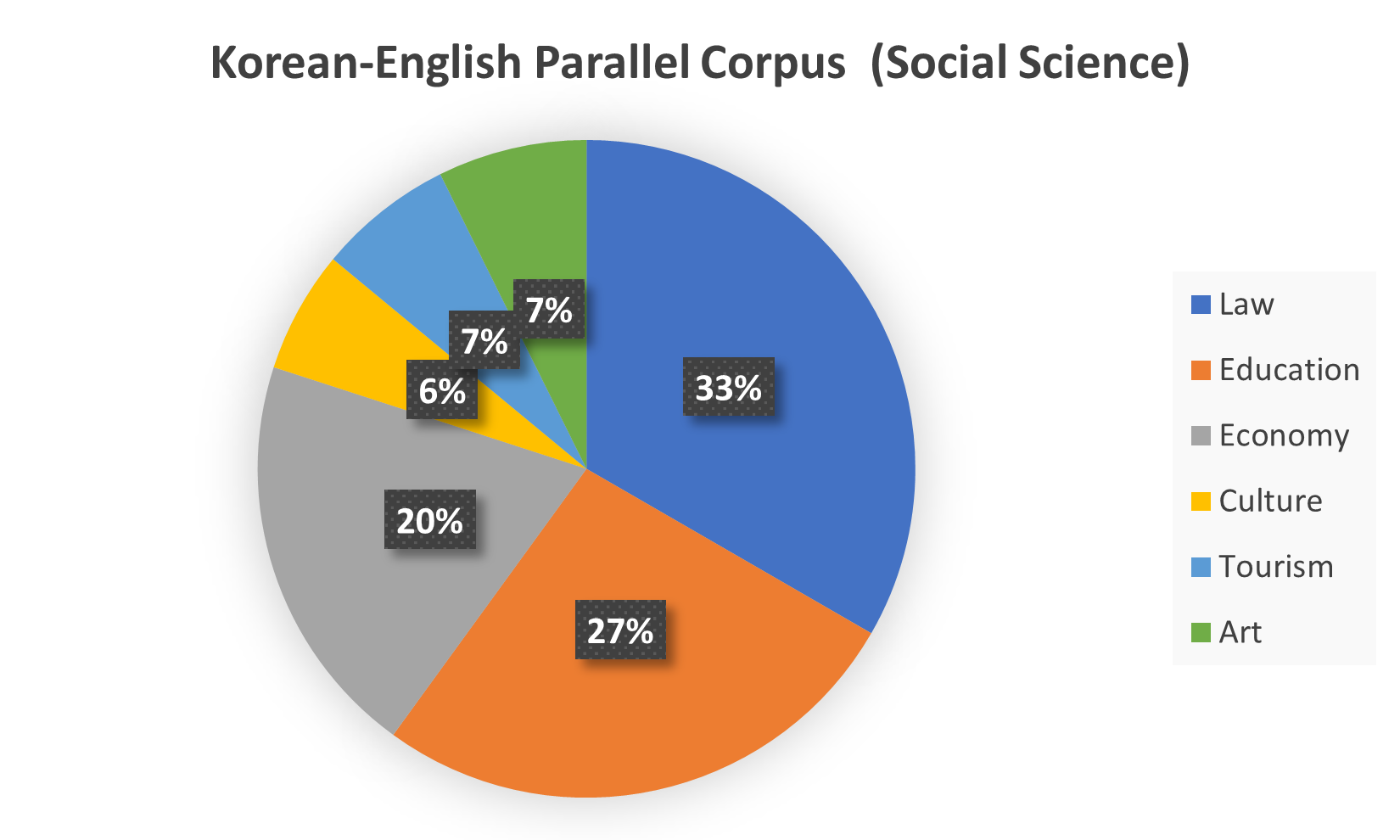}
  \caption{Data-domain statistics of the Korean-English Parallel Corpus (Social Science).}
  \label{fig:data_ko_en_social}
\end{figure}

The entire corpus comprises 1.5M sentence pairs, including 300K instances of economic data, 90K instances of cultural content, 100K instances of tourism content, 400K instances of education data, 500K instances of law data, and 110K instances of art domain content. The occupational ratio of each domain to the corresponding corpus is shown in Figure \ref{fig:data_ko_en_tech}. The data was revised by the specialists of their domain and translation experts. In this study, we partially leveraged 537K sentence pairs (law (37.2\%), economy (24.2\%), education (24.1\%), tourism (5.9\%), culture (4.5\%), art (4\%), and medical(0.08\%)) because all the data is yet to be released. 

\paragraph{Corpus Analysis}
The data was also split into the training and validation datasets, and the results of running LIWC on the training, validation, and the entire dataset are shown in the Table \ref{tab:3.4}. The difference between each linguistic feature of the training and validation datasets is mostly within 0.1\%, and as an exception, some features, such as `authentic' and `emotional tone,' have relatively sizable dissimilarities. This is as a result of the incidental blending of datasets with various domains, such as law, culture, and economy.

First, the morphological analysis is homogeneous to the results presented in Section \ref{sec:ko-en-para}, but negates serving as not and never have been used the most among other corpora. This result is contrary to the outcomes in the technical science corpus discussed in Section \ref{sec:ko-en-tech}, thereby suggesting that the frequency of plain and negative statements varies depending on the domain. In the case of pronouns, impersonal pronouns (`ipron') have been used four times more than personal pronouns (`ppron'), as shown in the results of Section \ref{sec:ko-en-para}, which is caused by the characteristics of the corpus in which the written sentences account for the description of most of the objects.

Considering syntactic characteristics, the number of analytic explanations is relatively high, thereby indicating that this corpus logically describes domains, such as economics, law, and education. There is a higher ratio of `Sixltr' compared to other corpora, thereby demonstrating the increased use of average long words. The low `WPS' also suggests that short sentences have been used. We also find that future focus (`focusfuture') accounts for the smallest percentage in the time orientations category. The reason behind this output is that there exist present state-oriented explanations rather than future predictions in the social and cultural corpora.

As a characteristic of the semantic perspective, the biological process category shows the lowest score compared to that of other corpora. Specifically, `body (body)' figures are within 0.2\% as a result of the nature of the social science domain, which is far from biology-related topics. Additionally, the cognitive process of human thinking is the highest compared to that of other corpora, with 1.5 to 1.8 times higher insight and cause. This confirms that `insight (insight)' and `cause (cause)' are attributes of written sentences in the social science domain in contrast to the technical science domain and specialized fields. The prevalence of `posemo' is twice as high as that of `negemo'. Subsequently, considering the phenomenon of gender bias, male-related pronouns are approximately twice as much as female-related pronouns, similar to other corpora. 

Similarly, because this corpus has a different ratio of domains from the dataset that would be completed at 1.5 million, as described in Section \ref{sec:ko-en-tech}, we can infer and analyze the linguistic features of each domain by comparing the results of the dataset that will be updated to 1.5 million.

\begin{table*}[!tbh]
\renewcommand{\arraystretch}{0.6}
\centering
\caption{\label{tab:3.4} LIWC results of Korean-English Parallel(Social Science) and Korean-English Parallel(Technology) Corpus.   }
\scalebox{0.7}{
\begin{tabular}{l|l|l|l|l|l|l|l}
\hline
                                        &              & \multicolumn{3}{l|}{Ko-En Parallel Corpus (Social Science)} & \multicolumn{3}{l}{Ko-En Parallel Corpus (Technology)} \\ \hline \hline
Categories                              & Features     & train               & valid              & total               & train                 & valid               & total                \\ \hline
\multirow{4}{*}{summary language variables}      & \textbf{Analytic}     & \textbf{97.38}               & \textbf{97.38}              & \textbf{97.38}               & \textbf{99 }                   & \textbf{99  }               & \textbf{99}                  \\ \cline{2-8} 
                                        & Clout        & 53.86               & 53.87              & 53.86               & 49.13                 & 49.17               & 49.13                \\ \cline{2-8} 
                                        & Authentic    & 23.99               & 24.26              & 24.02               & 16.69                 & 16.49               & 16.67                \\ \cline{2-8} 
                                        & Tone         & 47.66               & 47.91              & 47.69               & 34.02                 & 33.99               & 34.02                \\ \hline
\multirow{15}{*}{Linguistic dimensions} & \textbf{WC }          & \textbf{9,991,408}           & \textbf{1,250,115}         & \textbf{11,241,523}          & \textbf{13,621,209}            & \textbf{1,702,722}           & \textbf{15,323,931}           \\ \cline{2-8} 
                                        & WPS          & 20.77               & 20.79              & 20.77               & 17.87                 & 17.87               & 17.87                \\
\hline
\end{tabular}}
\end{table*}

\subsection{Korean-Chinese Parallel Corpus (Technology)}
\label{sec:ko-zh-tech}

\paragraph{Corpus Description}
AI Hub also provided the technology-domain specialized Korean-Chinese parallel corpus \footnote{\url{https://aihub.or.kr/aidata/30722}}. This corpus is the first publicly-released Korean-Chinese parallel corpus. To build this corpus, six companies, including Saltlux partners, Flitto, Evertran, Onasia \footnote{\url{https://on-asialang.com/}}, Yoon's information development company, and dmtlabs \footnote{\url{http://dmtlabs.co.kr/}} cooperated.

\begin{figure}[h!]
  \centering
  \includegraphics[scale=0.60]{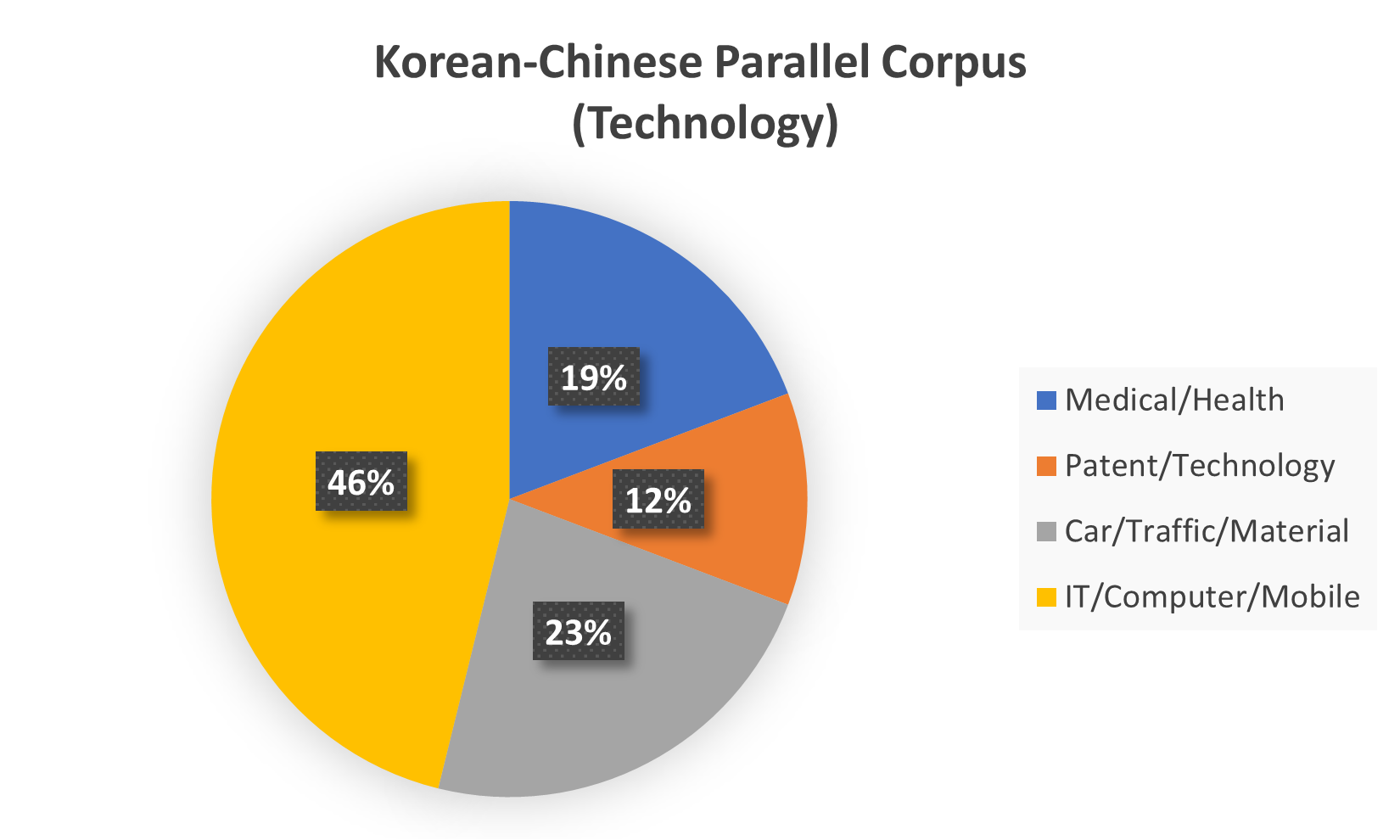}
  \caption{Data-domain statistics of the Korean-Chinese Parallel Corpus (Technology).}
  \label{fig:data_ko_zh_tech}
\end{figure}

The entire corpus comprises 1.3M sentence pairs, including 250K instances of medical/health data, 150K instances of patent/technology data, 300K instances of car/traffic/material data, and 600K instances of IT/computer/mobile-related content. Figure \ref{fig:data_ko_zh_tech} shows the ratio of each domain. This corpus is subdivided into the training and validation datasets, and Table \ref{tab:3.6} shows the LIWC analysis results of both datasets. 

Owing to the characteristics of the Chinese language, there exist differences between the training and validation datasets in count-based analyses, such as `WPS' and `Sixltr', but their severities are subtle. We inspect the linguistic characteristics of English and Chinese by comparing them between those of the technology-specialized Korean-English parallel corpus, which is analyzed in Section \ref{sec:ko-en-tech}.

\paragraph{Corpus Analysis}
In the morphological analysis, unlike Section~\ref{sec:ko-en-tech}'s result where `common verb' appears 1.7 times more than `auxverb,' this corpus shows the lowest difference between the two features at 1.45 times. Additionally, the results indicate that it rarely uses the negative representations, `quant' and `negate'. We establish that this corpus shows notable differences in pronoun analysis. Unlike Section~\ref{sec:ko-en-tech}, where personal pronouns were rarely used, personal pronouns were used approximately three times as often as non-personal pronouns, and among them, the `1st pers plural (we)' was the most common, and the `3rd pers pronouns' (i.e., 3rd pers singular and plural) were rarely used. 

Due to the nature of the data in technical field, declarative texts take large portion of the whole dataset, and thereby `semicolons (SemiC),' `Colons (Colon),' `Dashes (Dash),' and `QMark' were rarely used as demonstrated in Section \ref{sec:ko-en-tech}.

In the aspect of syntactic analysis, `analytic' and confidence, `Clout,' were the highest throughout all the English corpora we analyzed. As primary purpose of the data in technology-domain is to convey existing information that proposed priorly, the present tense is less focused than the past and future tenses.
 
Notably, `analytic' was 5.9\% lower than that presented in Section \ref{sec:ko-en-tech}'s `analytic,' and `WPS' and `Sixltr' were much higher. These results show that the length of sentences and words used in Chinese are longer than those used in English. Additionally, unlike most Korean-English parallel corpora, including those presented in Section \ref{sec:ko-en-tech}, `article' and `prep' are scarcely used, and the use of all tenses in the time orientations category with the exact weight is also a characteristic of Chinese.

In the punctuations category, the usage frequency of `colon' is similar to that presented in Section \ref{sec:ko-en-tech}'s results. This is a characteristic that explains the existence of multiple contents in one sentence. Additionally, `number,' which directly represents a number, was higher than `quant,' which represents a quantitative description. However, informal language markers were rarely used. It is noteworthy that `quotes,' which were hardly used in Korean-English parallel corpora, accounted for 12.7\%. This result suggests that the presence of many quotations in this corpus show the differences between the English corpus and the Chinese corpus.

Considering the semantic aspects, `work' and `leisure' have the highest ratios in the personal concerns category. In addition, the perception process, biological process, and cognitive process categories were higher than in other Korean-Chinese corpora, which is similar to the results presented in Section \ref{sec:ko-en-tech}. 

Through the qualitative inspection, we conclude that the semantic difference between the corpora with similar domains is identical, except morphological and syntactic distinction of each respective language.

Overall, in sentiment analysis, all outcomes in the affective process category, including emotional tone in the summary language variables category, are low. This is because, as discussed in Section \ref{sec:ko-en-tech}, it consists of a sentence-oriented corpus that describes knowledge and phenomena. The unusual thing is that in the corpus, `posemo' appeared approximately six times more than `negemo', which is similar to the results presented in Section \ref{sec:ko-en-tech}.

In the view of gender bias, most of the Korean-English parallel corpora analyzed so far had gender bias, but there was no gender bias in all the Chinese corpora.

\subsection{Korean-Chinese Parallel Corpus (Social Science)}
\label{sec:ko-zh-social}

\paragraph{Corpus Description}
Along with the technology-domain specialized corpus, AI Hub also released a social science-domain specialized Korean-Chinese parallel corpus \footnote{\url{https://aihub.or.kr/aidata/30721}}. To build this corpus, six companies, including Saltlux partners, Flitto, Evertran, Onasia, Yoon's information development company, and dmtlabs, cooperated.

\begin{figure}[h!]
  \centering
  \includegraphics[scale=0.60]{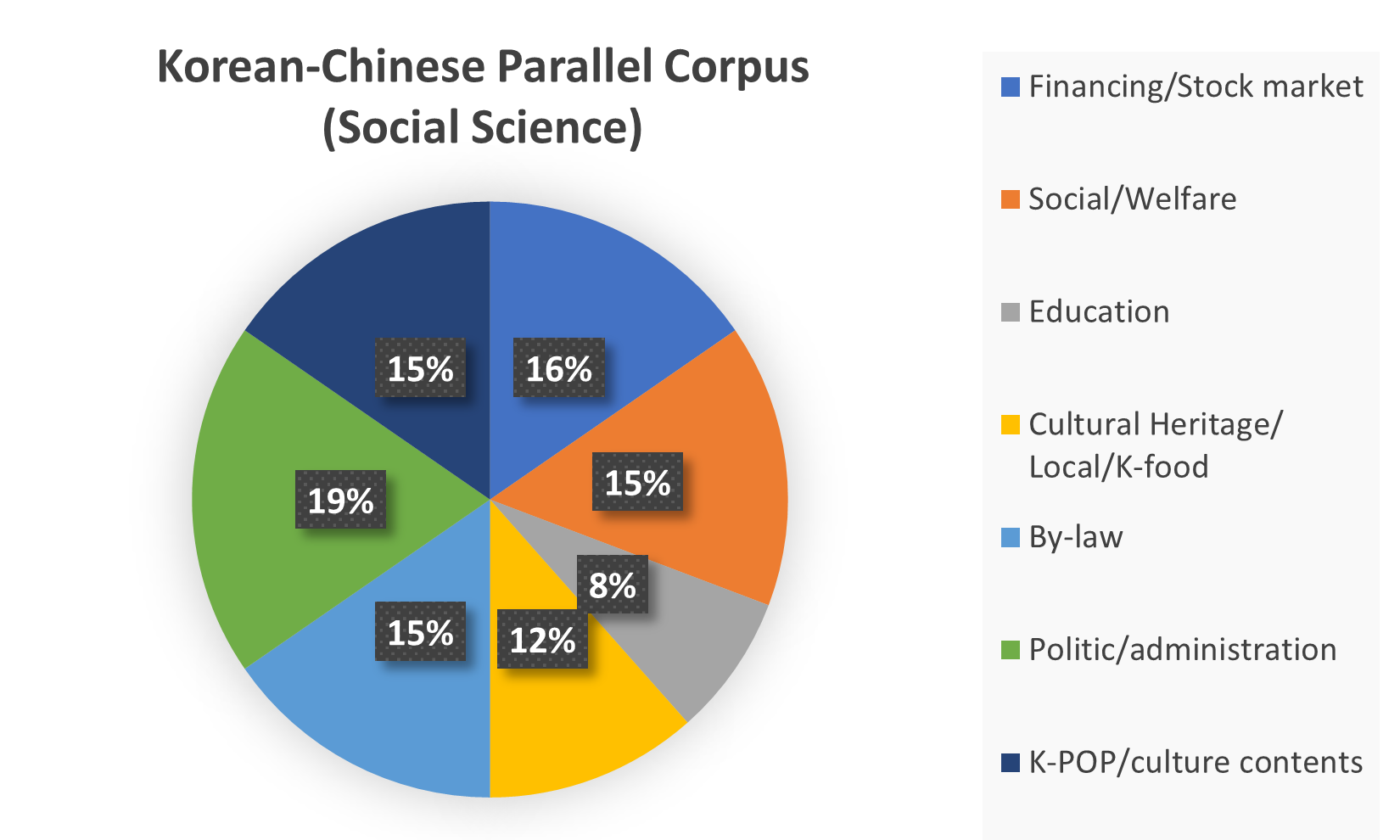}
  \caption{Data-domain statistics of the Korean-Chinese Parallel Corpus (Social Science).}
  \label{fig:data_ko_zh_social}
\end{figure}

The total amount of sentence pairs in the corpus is 1.3M, including 200K instances of financial/stock market contents, 200K instances of social/welfare domain data, 100K instances of education data, 150K instances of cultural heritage/local/K-food content, 250K by-law texts, 250K instances of political/administration data, and 200K instances of K-POP/culture content. The ratio of each domain to the entire corpus is shown in Figure \ref{fig:data_ko_zh_social}.

\paragraph{Corpus Analysis}
As shown in Table \ref{tab:3.6}, the overall characteristics of the training and validation datasets are almost identical. The overall analysis results are generally similar to the results presented in Section \ref{sec:ko-zh-tech}, except for a few aspects. 
The corresponding corpus showed a similar ratio of `conjunctions (conj),' `negations (negate),' `comparisons (compare),' and `interrogatives (interrog)' in the grammar category. Through the inspection of syntactic analysis, we established that the relatively frequent `preposition,' `comma,' `question mark (Qmark),' and `quote' are contained in each sentence. This shows that the length of each sentence is quite short, and the proportion of `questions' and `quotes' is relatively high. We can infer that descriptive methods that sequentially list various types of information have been commonly used.

We can point out a common feature with Section \ref{sec:ko-en-social} because personal pronouns are used three times as much as non-personal pronouns. However, the corresponding corpus has a distinguishable feature in that more first and second person singular pronouns are used more frequently than first person plural pronouns. These results show the attributes of the domain of the corresponding corpus, where the descriptions of social/culture/politics, which mainly focus on "I" and "You," are composed.

Furthermore in the corpus, the prevalence of `posemo' is higher than that of `negemo' and gender bias rarely exists. Later, by analyzing the linguistic and colloquial Chinese corpora in various fields, we verified whether it is a linguistic characteristic of Chinese or a special case occurring during descriptions in a specialized field.

\begin{table*}[!tbh]
\renewcommand{\arraystretch}{0.6}
\centering
\caption{\label{tab:3.6}LIWC results of Korean-Chinese(Zh) Parallel(Social Science) and Korean-Chinese(Zh) Parallel(Technology) Corpus.  }
\scalebox{0.7}{
\begin{tabular}{l|l|l|l|l|l|l|l}
\hline 
                                        &              & \multicolumn{3}{l|}{Ko-Zh Parallel corpus (Social Science)} & \multicolumn{3}{l}{Ko-Zh Parallel Corpus (Technology)} \\ \hline \hline
Categories                              & Features     & train                & valid             & total               & train                 & valid               & total                \\ \hline
\multirow{4}{*}{summary language variables}      & Analytic     & 93.25                & 93.24             & 93.25               & 93.15                 & 93.09               & 93.14                \\ \cline{2-8} 
                                        & Clout        & 50.44                & 50.49             & 50.45               & 51.73                 & 52.14               & 51.78                \\ \cline{2-8} 
                                        & Authentic    & 1                    & 1                 & 1                   & 1                     & 1                   & 1                    \\ \cline{2-8} 
                                        & \textbf{Tone}         & \textbf{26.97}                & \textbf{27.05 }            & \textbf{26.98 }              & \textbf{29.18  }               & \textbf{29.75 }              & \textbf{29.25}                \\ \hline
\multirow{15}{*}{Linguistic dimensions} & WC           & 3,907,897            & 482,441           & 4,390,338           & 3,954,651             & 512,873             & 4,467,524            \\ \cline{2-8} 
                                        & \textbf{WPS}          & \textbf{598.73}               & \textbf{724.39 }           & \textbf{612.54 }             & \textbf{2569.62}               & \textbf{2947.55}             & \textbf{2613.01}              \\
\hline
\end{tabular}}
\end{table*}

\subsection{Korean-Japanese Parallel Corpus}
AI Hub released the public Korean-Japanese parallel corpus \footnote{\url{https://aihub.or.kr/aidata/30723}} for the first time in Korea. Each sentence pair in the corpus is generated by translating Korean sentences from various domains into Japanese sentences using MT systems, after which it is revised by human experts. This corpus is not biased to a specific industrial domain and is constructed from a raw data source. Therefore, it is free from copyright problems. These attributes enable the corpus to be widely utilized for any NLP industrial services that deal with various domains. To build this corpus, six companies, including Saltlux partners, Flitto, Evertran, Onasia, Yoon's information development company, and dmtlabs, cooperated.

\begin{figure}[h!]
  \centering
  \includegraphics[scale=0.60]{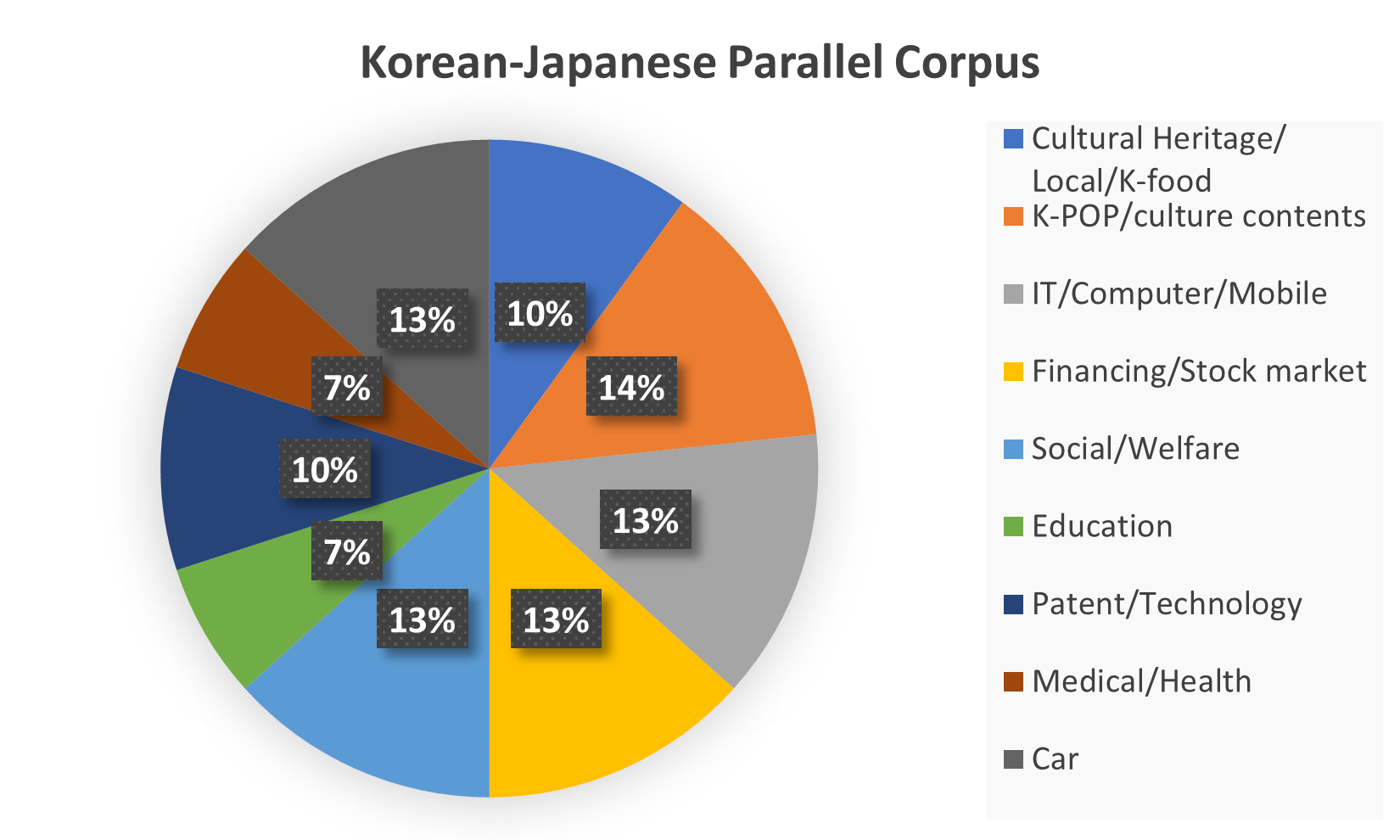}
  \caption{Data-domain statistics of the Korean-Japanese Parallel Corpus.}
  \label{fig:data_ko_jp}
\end{figure}

The entire corpus comprises 1.3M sentence pairs, including 150K instances of cultural heritage/local/K-food content, 200K instances of K-POP/culture content, 200K instances of IT/computer/mobile domain data, 200K instances of finance/stock market contents, 200K instances of social/welfare data, 100K instances of education data, 150K instances of patent/technology domain data, 100K instances of medical/health content, and 200K instances of car-related data. The ratio of each domain to the entire corpus is shown in Figure \ref{fig:data_ko_zh_social}. As the proper LIWC software has not been publicly released, we skipped the corpus analysis for the corresponding corpus.

\section{Experiments and Results}
\subsection{Dataset Details}
In this study, we utilize seven types of Korean parallel corpora released by AI Hub as training data for the experiments. We measure the total number of sentences for each corpus, the minimum, maximum, and average length for each word, and the character unit. 
Statistics for the seven newly released parallel corpora by AI Hub are listed in Table \ref{result-HUB}. In the case of the social science and technology fields of the Korean-English parallel corpus, only 470K and 690K instances of data were released owing to unintentional circumstances of the organizers. We leverage the official training and validation datasets for training and evaluation. Before training, to test the performance of the MT system, we use a partially separated 3K instance of the training set as a test set.
Performance of each NMT model in our experiments is measured by the BLEU score \cite{papineni2002bleu}, which is the common metric in NMT field, and for the precise evaluation, we adopted Jieba \footnote{\url{https://github.com/fxsjy/jieba}} and MeCab \footnote{\url{https://github.com/taku910/mecab}} as tokenizers of Chinese and Japanese output sequence.

\begin{table*}[!tbh]
\renewcommand{\arraystretch}{1.0}
\centering
\caption{\label{result-HUB} Summary of overall AIHUB datasets. For statistics on tokens, we denote NA because there are no spaces in Japanese and Chinese.}
\scalebox{0.53}{
\begin{tabular}{lccccccccc}
\toprule
\multicolumn{3}{l|}{} & \multicolumn{1}{c|}{\textbf{\# of Sents}} & \multicolumn{1}{c|}{\textbf{\# of min toks}}  & \multicolumn{1}{c|}{\textbf{\# of max toks}} & \multicolumn{1}{c|}{\textbf{Avg toks per S}}  & \multicolumn{1}{c|}{\textbf{\# of min chars}}  & \multicolumn{1}{c|}{\textbf{\# of max chars}} & \multicolumn{1}{c}{\textbf{Avg chars per S}}\\ \midrule \hline

\multicolumn{1}{l|}{\multirow{6}{*}{\textbf{Korean-English Parallel Corpus}}} & \multicolumn{1}{c|}{\multirow{2}{*}{\textbf{Train}}} & \multicolumn{1}{c|}{\textbf{KO}} & \multicolumn{1}{c|}{1,399,116} & \multicolumn{1}{c|}{1} & \multicolumn{1}{c|}{78} & \multicolumn{1}{c|}{12.97}  & \multicolumn{1}{c|}{4}  & \multicolumn{1}{c|}{359}  & \multicolumn{1}{c}{55.07} \\ \cline{3-10}
\multicolumn{1}{c|}{} & \multicolumn{1}{c|}{} & \multicolumn{1}{c|}{\textbf{EN}} & \multicolumn{1}{c|}{1,399,116} & \multicolumn{1}{c|}{2} & \multicolumn{1}{c|}{180} & \multicolumn{1}{c|}{22.57}  & \multicolumn{1}{c|}{10}  & \multicolumn{1}{c|}{999}  & \multicolumn{1}{c}{135.24} \\ \cline{2-10}
\multicolumn{1}{c|}{} & \multicolumn{1}{c|}{\multirow{2}{*}{\textbf{Valid}}} & \multicolumn{1}{c|}{\textbf{KO}} & \multicolumn{1}{c|}{200,302} & \multicolumn{1}{c|}{2} & \multicolumn{1}{c|}{32} & \multicolumn{1}{c|}{17.46}  & \multicolumn{1}{c|}{9}  & \multicolumn{1}{c|}{220}  & \multicolumn{1}{c}{75.36} \\ \cline{3-10}
\multicolumn{1}{c|}{} & \multicolumn{1}{c|}{} & \multicolumn{1}{c|}{\textbf{EN}} & \multicolumn{1}{c|}{200,302} & \multicolumn{1}{c|}{3} & \multicolumn{1}{c|}{110} & \multicolumn{1}{c|}{30.73}  & \multicolumn{1}{c|}{14}  & \multicolumn{1}{c|}{706}  & \multicolumn{1}{c}{187.89} \\
\bottomrule
\end{tabular}}
\end{table*}

\subsection{Models Detail}
For verifying the quality of dataset provided by AI Hub, we constructed transformer based NMT model \cite{vaswani2017attention} that trained with each dataset. Transformer is an auto-regressive model structure which comprises encoder-decoder architecture, and is widely utilized in many NLP research fields, including NMT, in achieving SOTA performance. Corresponding model refrains recurrence and constructs its encoder-decoder model architecture by mainly applying attention structure. This enables considerable reduction of required training time by allowing significantly more parallelization in training process. Attention based model structure of transformer also can relieve long term dependency problem of RNN and LSTM \cite{hochreiter1997long}. Output results of attention structure can be described as equation(\ref{eq:attention}).

\begin{align} \label{eq:attention}
\resizebox{0.88\columnwidth}{!}{${
\text { Attention }(q, k, v)=\operatorname{Softmax}\left( \frac{(W_{q} q) (W_{k} k)^{T}}{\sqrt{d_{k}}}\right) (W_{v} v)
}$}
\end{align}

In equation (\ref{eq:attention}), $W_{q}$, $W_{k}$ and $W_{v}$ refers to trainable parameters. Attention structure takes three input; query, key and value, which is denoted as $q$, $k$ and $v$, respectively. Through this structure, transformer can obtain the relational information between input sentence and generating sentence. In such cases, the embedding obtained from input sentence is fed to the attention structure as $q$ and $k$, and the embedding from the generating sentence is regarded as $v$. Attention structure is also leveraged to obtain the bidirectional contextual information of input sentence and generating sentence, through self-attention mechanism which takes identical embedding value as $q$, $k$, and $v$ simultaneously. 

We construct transformer NMT model trained with each AI Hub dataset. We regard the performance of NMT model as the quality of parallel corpus, by controlling all the training conditions of our experiments to be identical, except the training dataset. Training objective of transformer based NMT model $\theta$ that trained with parallel corpus $P$ can be described as equation (\ref{eq:training_objective}). 

\begin{equation} \label{eq:training_objective}
\resizebox{0.88\columnwidth}{!}{${
    \max\limits_{\theta}  \ \cfrac{1}{\Vert D \Vert} \sum_{(X, Y) \in D} \log \left[ \prod_{i=1}^n P(y_{i} \mid X, y_{t<i}, \theta) \right]
}$}
\end{equation}

Overall process is similar to the training of sequence to sequence \cite{sutskever2014sequence} based MT model. In (\ref{eq:training_objective}), $X$ and $Y$ indicate source and target sentence in $P$, respectively. Target sentence $Y$ comprises total $m$ tokens, which are denoted as $\{y_i\}_{1}^{n}$, and through this training process, corresponding model is trained to generate $Y$ auto-regressively.

In our training process, we used adam optimizer with noam decay, and all the batch size is set to be 4096. The transformer NMT model in our experiments consists of six encoder and decoder layers with six attention blocks and eight attention heads, which dimensionality and embedding size is 512. 

For the pre-processing of our training data, we utilized sentencepiece \cite{kudo2018sentencepiece} subword tokenization method, with 32,000 vocab size. We extracted 5,000 and 3,000 samples randomly from training data for the validation and test set, respectively. The performance evaluation of all the translation results are proceeded with BLEU score by leveraging multi-bleu.perl script\footnote{\url{https://github.com/moses-smt/mosesdecoder/blob/master/scripts/generic/multi-bleu.perl}} given by Moses.

\subsection{Main Results}
\begin{table*}[!tbh]
\renewcommand{\arraystretch}{1.0}
\centering
\caption{\label{tab:result} Experimental results for seven datasets and three language pairs published by AI Hub.}
\scalebox{0.8}{
\begin{tabular}{l|c|c}
\toprule
Corpus                                                                   & Language & BLEU  \\ \midrule \hline
\multirow{2}{*}{Korean-English Parallel corpus}                    & KO-EN    & 28.36 \\ \cline{2-3} 
                                                                   & EN-KO    & 13.53 \\ \hline
\multirow{2}{*}{Korean-English Parallel corpus (Social Science)}   & KO-EN    & 45.64 \\ \cline{2-3} 
                                                                   & EN-KO    & 17.71 \\ \hline
\multirow{2}{*}{Korean-English Parallel corpus (Technology)}       & KO-EN    & 63.88 \\ \cline{2-3} 
                                                                   & EN-KO    & 39.17 \\ \hline
\multirow{2}{*}{Korean-English Domain-specialized Parallel corpus} & KO-EN    & 51.88 \\ \cline{2-3} 
                                                                   & EN-KO    & 21.99 \\ \hline
\multirow{2}{*}{Korean-Japanese Parallel corpus}                   & KO-JA    & 68.88 \\ \cline{2-3} 
                                                                   & JA-KO    & 49.05 \\ \hline
\multirow{2}{*}{Korean-Chinese Parallel corpus (Social Science)}   & KO-ZH    & 48.74 \\ \cline{2-3} 
                                                                   & ZH-KO    & 25.16 \\ \hline
\multirow{2}{*}{Korean-Chinese Parallel corpus (Technology)}       & KO-ZH    & 46.70 \\ \cline{2-3} 
                                                                   & ZH-KO    & 25.75 \\ \bottomrule
\end{tabular}} 

\end{table*} 

\paragraph{Performance analysis}
The baseline results of the seven AI Hub parallel corpora are listed in Table \ref{tab:result}. The experiment showed a BLEU score of 28.36 for the Korean-English NMT model trained using Korean-English parallel corpora. In the case of the NMT model trained using the Korean-English parallel corpus (technology), the performance showed a BLEU score of over 50, which shows that the words and expressions in a specific domain appear quite repeatedly. For the NMT models based on other fields, the Korean-English parallel corpus (social science) and the Korean-English domain-specialized parallel corpus also demonstrated high performance of 45.64 and 51.88, respectively, in similar contexts. 

Considering the significant performance gap between domain and general corpora, we can point out probable limitation of corpus construction. Although the performance of all domain corpora is overwhelmingly higher than general corpora, this result does not guarantee that the NMT model is well operating because we randomly extract test set within the training set. Corpora built on the basis of a particular domain typically have significant overlap parts with other sentences within such corpora, but there still exist many different expressions and words present in the field. This can cause difficulties in translating other various expressions. Therefore, our experimental results show that corpus generators should include much more diverse expressions especially in specific domains given that a well-constructed corpora makes a model smarter.

The Korean to Japanese NMT model based on the Korean-Japanese parallel corpus scored 68.88. The BLEU scores of the NMT models trained using the Korean-Chinese parallel corpus (Social Science) and the Korean-Chinese parallel corpus (Technology) are 48.74 and 46.70, respectively.

\paragraph{Language direction analysis}
As a result of conducting both Korean to English translation and the opposite based on four Korean-English parallel corpora, the gap of the experimental results for translating Korean to English compared to those of the opposite case differ significantly, with scores from 14.83 to 29.89. 

This can be interpreted in terms of data construction. Using parallel corpora built by translating sentences from one language to another, translation results can be awkward when training the model in the opposite direction. Thus, a reasonable construction process for training direction-robust NMT models involves building a parallel corpus by constructing about half as the source language and the other half as the target language and translating each. In other words, given the significant differences in performance when changing the direction of translation, it is highly likely that the translation was carried out using only a monolingual corpus, which consists of a source language without considering the opposite direction. Similarly, in the case of Korean-Japanese and Korean-Chinese models, the performance in the opposite direction was significantly reduced. These aspects should be considered when building parallel datasets in the future.

In this paper, such a problem is defined as ``data imbalance''~\cite{cai2015challenges,park2021study}, and the problem must be solved when constructing data in the future . As for high-quality data, it is important that various elements are ultimately built in a balanced manner, and we conducted further analysis in this respect.

\begin{figure*}[th!]
  \centering
  \includegraphics[scale=0.25]{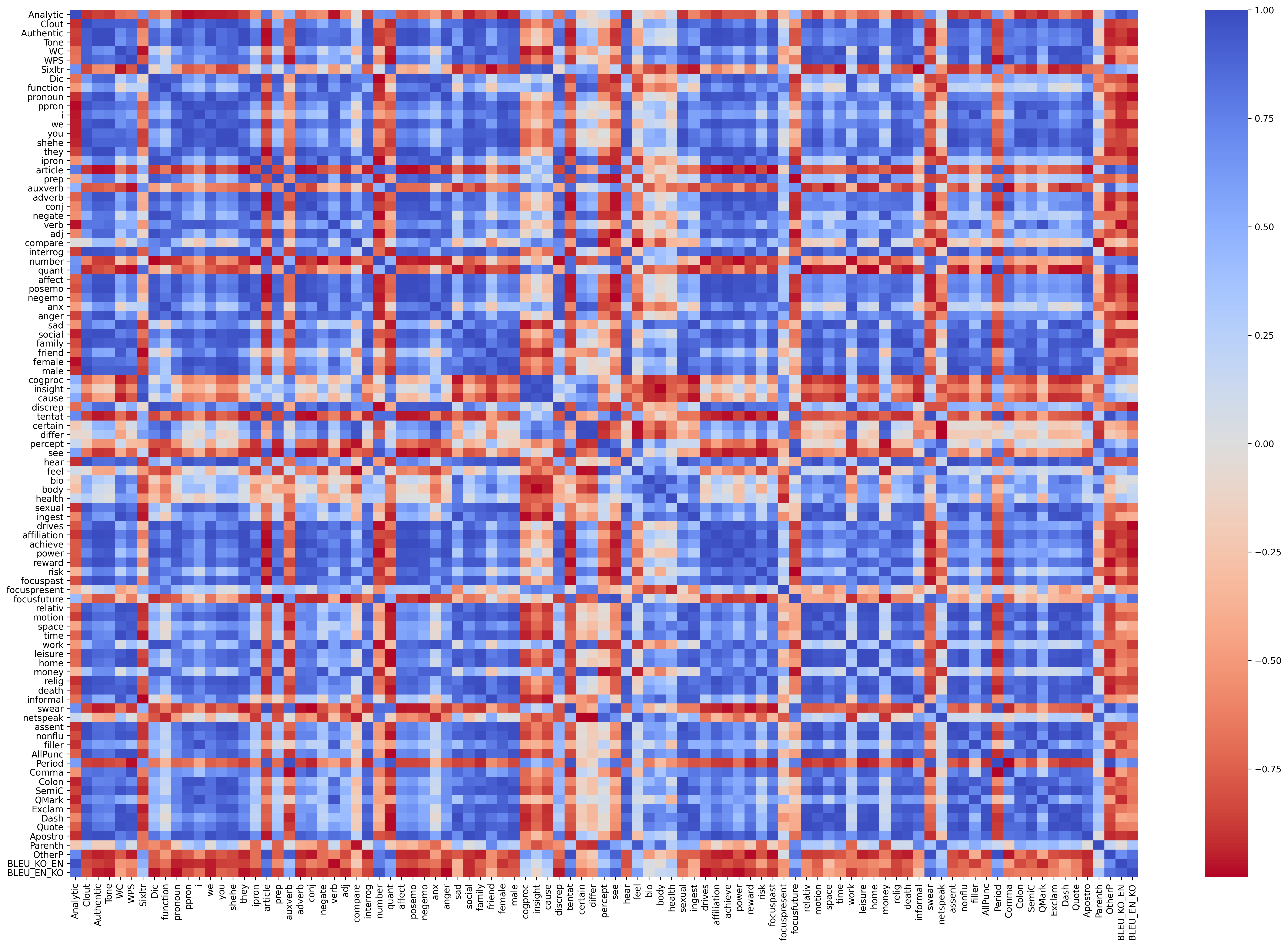}
  \caption{Results of the correlation between LIWC features and BLEU score (KO-EN). The blue-colored indicates a positive correlation while red-colored indicates a negative correlation.}
  \label{fig:liwc_corr}
\end{figure*}

\paragraph{Correlation Analysis between LIWC and BLEU score} We analyzed a correlation between LIWC features and BLEU score for observing the connection between them. We employed BLEU score derived from the Korean-English corpus results in Table~\ref{tab:result} and used only LIWC features of the Korean-English case to do this. As shown in Figure~\ref{fig:liwc_corr}, we calculated a Pearson correlation~\cite{benesty2009pearson} joining all the features coming from LIWC, BLEU score (KO-EN), and BLUE (EN-KO).

\begin{figure*}[th!]
  \centering
  \includegraphics[scale=0.5]{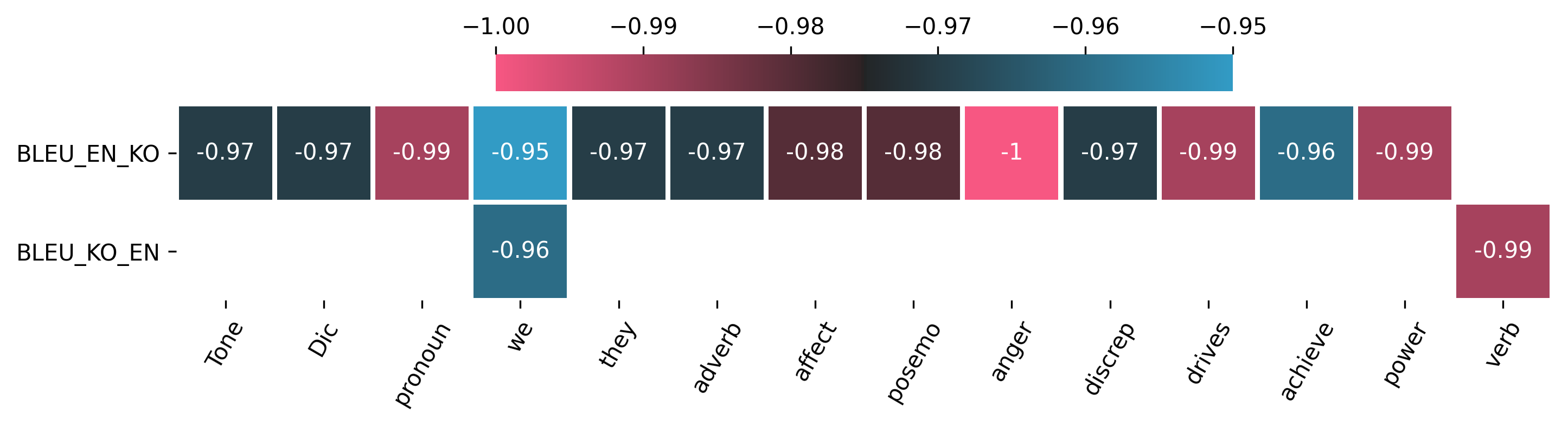}
  \caption{Negative correlation ($r<0$) results of the important factors between BLEU score (KO-EN) and LIWC features. The empty cells (white) indicate cut-offed due to the positive value. Note that this result is statistically significant as $p<0.05$.}
  \label{fig:liwc_impo_corr}
\end{figure*}

We can infer following result with Table \ref{fig:liwc_corr}. First, the overall tendency of correlation within LIWC features is mostly not different from analysis in Section \ref{sec:LIWC}. For example, Analytic and sentimental levels show a negative correlation. This is a unified result since Analytic indicates whether emotions are excluded and tone of the text is logical. In other words, the results give validity to the LIWC analysis.

Secondly, we show that correlations between LIWC features and BLEU score is highly negative. It can be said that training data is good when it has balance in tone, length, gender and so on. However, there are many things to improve in AI Hub such as word count, punctuation usage, sentimental analysis. It is because there are numerous negative effects in terms of data imbalance. This suggests in which direction we should build data and informs us that the performances can be improved through data cleaning such as PCF \cite{koehn2020findings, park-etal-2021-bts}.

Additionally, we distilled the features by the case of statistically significant negative correlation in Figure~\ref{fig:liwc_impo_corr}. The most striking result to emerge from the BLEU score between En-Ko and Ko-En is that \textit{English-Korean translation affects the negative effect on BLEU score by those features, and many features need to filter than Korean-English translation.} This result suggests further research on which factors are considered to remove during the data filtering process. Also, our findings are supported by BLEU score about showing the lower score in EN-KO than KO-EN in terms of data imbalance as shown in Table~\ref{tab:result}.

Finally, this paper figured out the association between LIWC and BLEU score in terms of data filtering. It may assist to make guidelines for building datasets later.

\section{Discussion and Positive Impact of this study} 
This paper conducted in-depth analyses on various parallel corpora published by AIHUB. Structural components that directly determine the quality of each corpus were closely investigated through the LIWC, and the actual usability of each corpus was quantitatively evaluated through the NMT model trained by the corresponding corpus. Through these, we have posed a positive impact on the machine translation research fields and figured out the desirable direction of data construction. Specifically, main contributions of our paper can be described as follows:

First, to the best of our knowledge, for the first time, we performed quantitative and qualitative in-depth analyses on AIHUB data. We adopted LIWC as an investigation tool for parallel corpus, and derived various meaningful information (\emph{e.g. quality of parallel corpus}) by newly interpreting each component obtained from LIWC. As LIWC was generally used in psychological research, various aspects of corpus analysis were possible, such as morphological analysis, Syntactic analysis, and so on. It can be confirmed that the results were suitable for the features of each corpus in most cases. For example, in Section \ref{sec:ko-en-para}, informal language markers such as swear word and filler rather used although they rarely used in other corpus. It is because the corpus includes dialogues and spoken words. Additionally, the result of Word Count and Commas in Section \ref{sec:ko-en-domain} showed that Domain-Specialized Parallel Corpus tends to explain terminology in long sentences with commas. In Section \ref{sec:ko-en-tech}, the Emotional tone of the technology corpus was relatively low in order to concisely explicate the terminology, not a description of emotions. Since there are many texts with the economy as a topic in Section \ref{sec:ko-en-social}, money of personal interest topic has the highest rate than other corpora. Furthermore, word per sentence in Section \ref{sec:ko-zh-tech} and \ref{sec:ko-zh-social} was the longest due to the characteristics of Chinese, which rarely uses spaces. Away from model-centric machine translation studies, this paper encourages data-centric research differently. This can have a positive impact on the NMT research field by presenting a new perspective.

Second, We pointed out the problems of the data construction process by revealing that there was a significant discrepancy in performance between English-Korean and Korean-English NMT model trained by the identical parallel corpus. It can be inferred that it is caused by the improper construction strategy. When constructing a parallel corpus that comprises certain two languages (\emph{i.e. Korean and English}), it is desirable to construct a balanced corpus by translating a half of the translation into the first language based on the second language and the remaining half of the translation into the second language based on the first language. Through the empirical analysis, we point out that this aspect may underestimated.

For the last, we revealed that several important factors that determine the quality of corpus. In Section \ref{sec:ko-en-social}, we can infer that domain uniformity is neglected because it contains medical text in social science corpus. The gender bias, which had a major influence on the quality of corpus, was also overlooked in several corpora, especially in Section \ref{sec:ko-en-domain} and \ref{sec:ko-en-social}, there was a double gender bias. Additionally, we proposed that subject omission and cross-reference resolution problems should be further considered for ensuring the high quality data.

Eventually, this paper clearly analyzed the strengths and weaknesses of the existing AIHUB data and provided insight into the future direction of data construction. 

In general, in the case of data filtering, mathematical and modeling approaches are taken \cite{koehn2020findings, zhang2020parallel}. Those approaches also can be reflected in our future direction of data construction. There are LIWC analysis studies on the text~\cite{pope2016analysis,fast2016empath}. Previous research could be one of our options to enhance our approach. A topic-related approach is useful to filter unrelated topics by LIWC analysis~\cite{fast2016empath}.

There are still limitations in the language pairs provided by LIWC. LIWC supports diverse languages include Arabic, Chinese, Dutch, English, French, German, Italian, Portuguese, Russian, Serbian, Spanish, and Turkish. They are being used in psychological or linguistic research in various countries \cite{garzon2020validacion, paixao2020fake}. However, LIWC is not supported in low-resource languages such as Korean or Japanese since its dictionary has not been created. In the case of Korean, K-LIWC\cite{Lee2004korean} was once available and there are some studies using it \cite{Kim2016k}. However, it could not analyze Korean corpora since K-LIWC is currently unavailable.

\section{Conclusions}
In this work, we proceeded with a quality evaluation of all the Korean-related parallel corpus, released by AI Hub. 
For the model-centric performance validation, we constructed a transformer based NMT model trained with each parallel corpus. Through quantitative and qualitative analysis of these NMT models, we point out some probable limitations on constructing corpora. First, for learning NMT model well in specific field, the domain corpora should contain various words and expressions in consideration of the excessive performance difference between domain and general corpora. Second, given the significant performance gap in terms of language direction, half of the parallel data to be built must be configured in the source language and the other half in the target language and then translated respectively.

Away from the model-centric analysis, we encouraged data-centric research through LIWC analysis. We figured out the association between LIWC and model performance in terms of data filtering. Through this analysis, we suggested the direction of further work to improve model performance.
The national level re-examination of the various standards and building processes should be made for the encouragement of AI data construction researches. In the future, we plan to investigate efficient beam search strategies and new decoding methods by utilizing these AI Hub data. Also, to more accurately measure the model performance, we plan to build an official Korean-English test set.

\bibliography{anthology,custom}
\bibliographystyle{acl_natbib}

\end{document}